\documentclass[10pt,twocolumn,letterpaper]{article}
\usepackage[pagenumbers]{wacv}
\usepackage{graphicx}
\usepackage{amsmath}
\usepackage{amssymb}
\usepackage{booktabs}
\usepackage{multirow}
\usepackage{array}
\usepackage{xspace}
\usepackage{soul}

\soulregister\ref7
\soulregister\pageref7
\soulregister\cite7
\soulregister\name7

\DeclareMathOperator*{\argmax}{arg\,max}

\newcommand{\ci}[1]{{\scriptsize[#1]}}

\newcommand{\pendingnum}{\textcolor{red}{\textbf{TO RUN}}}

\newcommand{\name}{GeoDIVE\xspace}

\usepackage{algorithm}
\usepackage{algpseudocode}

\newcommand{\vnum}[1]{\if\relax\detokenize{#1}\relax\pendingnum\else#1\fi}
\renewcommand{\hl}[1]{{#1}}
\definecolor{wacvblue}{rgb}{0.21,0.49,0.74}
\usepackage[pagebackref,breaklinks,colorlinks,allcolors=wacvblue]{hyperref}
\hypersetup{pdftitle={GeoDIVE: Geospatial Descent Informed by Value Estimation},pdfauthor={Hamza Iqbal, Dan Cher, Eric Xing, Pranavi Kolouju, Abby Stylianou, Robert Pless, Nathan Jacobs}}
\def\wacvPaperID{1547}
\def\confName{WACV}
\def\confYear{2027}
\title{GeoDIVE: Geospatial Descent Informed by Value Estimation}
\author{Hamza Iqbal$^{1}$ \quad Dan Cher$^{1}$ \quad Eric Xing$^{1}$ \quad Pranavi Kolouju$^{2}$\\
Abby Stylianou$^{2}$ \quad Robert Pless$^{3}$ \quad Nathan Jacobs$^{1}$\\
$^{1}$Washington University in St. Louis, USA\\
$^{2}$Saint Louis University, USA \quad $^{3}$The George Washington University, USA
}

\begin{document}
\maketitle
\thispagestyle{empty}
\AddToShipoutPicture*{\AtPageLowerLeft{\put(54,24){\parbox[b]{504pt}{\fontsize{7}{8}\selectfont
\textcopyright\ 2027 IEEE. Personal use of this material is permitted. Permission from IEEE must be obtained for all other uses, in any current or future media, including reprinting/republishing this material for advertising or promotional purposes, creating new collective works, for resale or redistribution to servers or lists, or reuse of any copyrighted component of this work in other works.
}}}}

\begin{abstract}
    Finding specific objects, structures, and spatial patterns supports critical work in agriculture, infrastructure planning, environmental monitoring, and disaster response. These features, however, are often dispersed across very large regions and span dramatically different spatial scales, which defines a search space that is costly to embed, maintain, and search naively.
    We introduce GeoDIVE, a model that treats a multiresolution satellite image gallery as a forest and uses best-first traversal to determine which tiles to encode. At each visited node, a lightweight model then predicts an optimistic estimate of the best query similarity hidden among its descendants using the query and subtree representations. 
    In evaluations with image and open-vocabulary text queries, GeoDIVE matches exhaustive search at 10\% of the encoding cost for image queries and leads all tested baselines at every budget in both modalities. To support future evaluation of multiscale geospatial search, we also release object-level image and text queries for Git-10M, constructed from Microsoft building footprints and OpenStreetMap polygons with VLM-generated captions. Source code and models will be made publicly available.
\end{abstract}

\section{Introduction}
\label{sec:intro}

\begin{figure*}[t]
\centering
\includegraphics[width=1\textwidth]{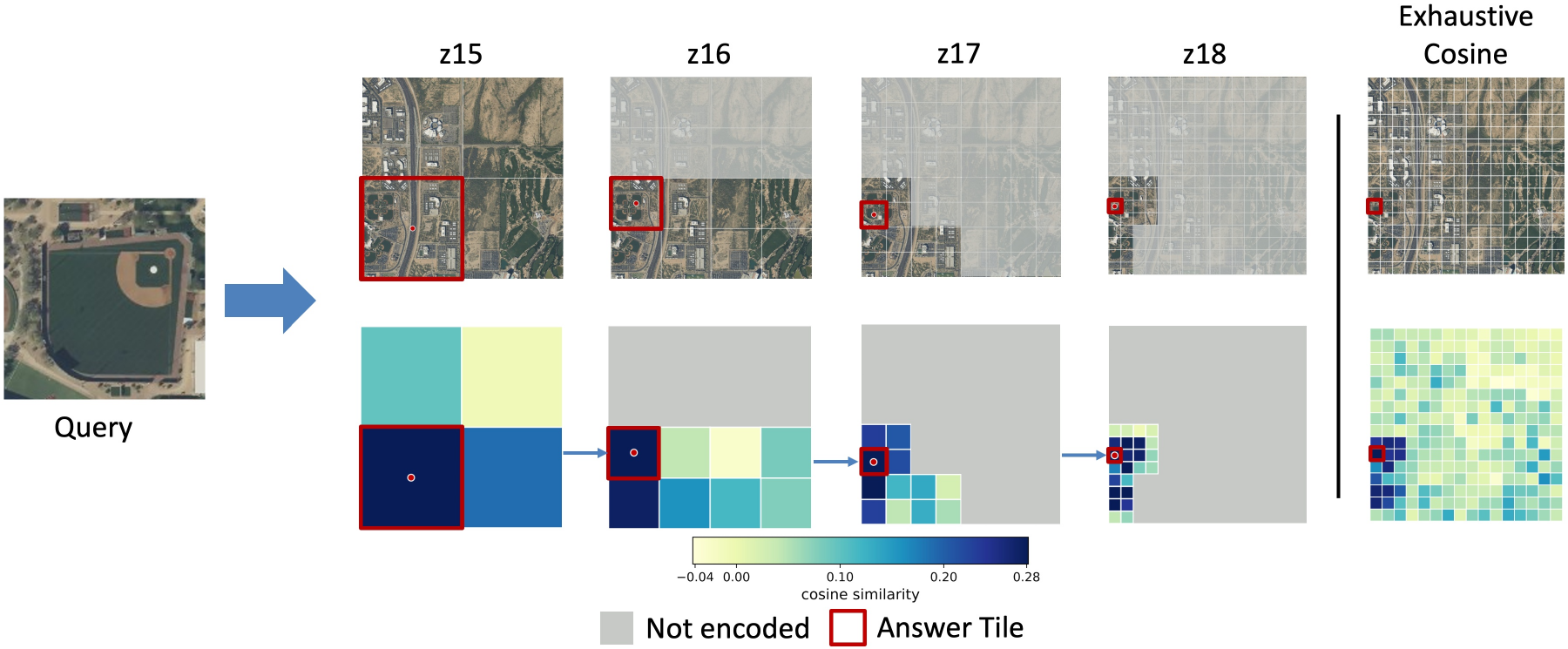}
\caption{\textbf{Encode everything, or descend.} A real descent for the query \emph{baseball field} over one gallery site, zoom level 15 (z15) to z18: the imagery (top) and the descent's similarity map of the same levels (bottom) under a 2\% encode budget. Colored cells were encoded and scored. Grey cells were never passed through the encoder. The red outline marks the answer tile and arrows follow the descent. Exhaustive cosine (right) must encode every leaf of the site to score the same target.}
\label{fig:descent-trace}
\end{figure*}

Visual search over a gallery typically encodes every image once and ranks queries against the resulting index. Some domains, like satellite imagery, hold billion-scale multi-resolution galleries that refresh continually. For instance, covering the contiguous USA at sub-meter resolution requires more than \textit{550 million} tiles. Additionally, targets of interest span multiple spatial scales, e.g., a vehicle cluster, a building, a stadium, and an airport become visually useful at very different zoom levels. A general-purpose framework for geospatial search must therefore cover a multi-resolution satellite image \textit{pyramid}.

Existing literature in embedding-based search acceleration focuses largely on accelerating a predetermined set of unrelated queries. Literature in approximate nearest-neighbor (ANN) search~\cite{jegou2011pq,malkov2020hnsw,guo2020scann} reduce the computational cost of individual queries in exchange for accuracy. Embedding sparsification~\cite{wen2025beyond, guo2026csrv} or truncation~\cite{kusupati2022matryoshka} methods also fundamentally operate on the per-embedding level, which is not sufficient to make exhaustive search tractable. These methods also assume an index that is built once, which every refresh of the gallery invalidates.

Intuitively, the prohibitive cost of geospatial search comes from the size of the image pyramid, which grows exponentially with each added zoom level. The content of those finer tiles, however, is partially predictable from the coarser imagery above them. A coarse tile of open farmland cannot conceal a school, so a search can prune its entire subtree without encoding a single child, while a dense city block warrants descent before any school is visible. We introduce GeoDIVE, an algorithm for geospatial search informed by probabilistic value estimation. A coarse tile matched against the query scores only what is visible at that resolution, and a school hiding in one of its children barely moves that score. GeoDIVE instead learns an optimistic estimate of the best match hidden below each tile. The search descends where that estimate is high and never encodes the rest. We summarize our contributions as follows:

\begin{itemize}\setlength{\itemsep}{1pt}
\item \textbf{Lazy encoding as search.} 
We reframe visual search over satellite imagery as a hierarchical search task. We optimize traversal over this forest of multi-zoom imagery via best-first search that matches the performance of an exhaustive dense search with significantly lower cost.
\item \textbf{Optimistic bound estimation.} We demonstrate that learning an optimistic estimate of subtree value is a strong heuristic to aggressively prune candidates during search.
\item \textbf{Efficient Late Interaction.} Our formulation of hierarchical traversal produces patch tokens for every tile it visits, enabling efficient multi-vector reranking without additional overhead cost. We demonstrate that this multi-vector scoring outperforms or matches exhaustive search on every reported metric while still only encoding 6\% of the gallery.
\end{itemize}

\section{Related Work}
\label{sec:related_work}

\paragraph{Tree retrieval and learned priorities.}
Many datasets are naturally hierarchical, from product catalogs to satellite image pyramids. Structuring them as trees has proven an efficient way to search them, and a body of work learns the traversal itself. Tree-based Deep Models select from large predefined label spaces~\cite{zhu2018tdm}, and later work scales them with beam-search inference~\cite{zhuo2020otm}. Efficient subwindow search constructs admissible upper bounds on subwindow value to avoid exhaustive spatial evaluation~\cite{lampert2008ess}. That guarantee makes its branch-and-bound exact, whereas our construction is learned from the data. Learned routing and early termination guide traversal of trees and similarity graphs~\cite{li2020earlytermination,baranchuk2019graphrouting,he2014learningtosearch}. GeoToken casts geolocalization as next-token prediction over the hierarchical S2 grid with beam-search inference (quadtree descent as sequence decoding) but reports no efficiency curves, a gap we address~\cite{ghasemi2025geotokenicdm}.
In settings where galleries are massive and continually refreshed, encoding carries a high per-query cost. We therefore learn a priority that judges each subtree from the coarse tile above it and encode only where the search descends.

\paragraph{Adaptive high-resolution processing.}
Attention sampling~\cite{katharopoulos2019attention} and FOVEA~\cite{thavamani2021fovea} concentrate pixels through learned or prior-guided resampling. PatchDrop~\cite{uzkent2020patchdrop} decides where paired high-resolution imagery is worth acquiring, and visual active search learns a meta-policy for inspecting geospatial cells under a local-search budget~\cite{sarkar2022vas}. AdaGlimpse chooses arbitrary glimpse locations and scales~\cite{pardyl2024adaglimpse}, and Dynamic Zoom-In Networks regress the detection gain of processing a region at higher resolution~\cite{gao2018dzn}. QuadTree Attention prunes token-level attention coarse-to-fine within a transformer~\cite{tang2022quadtree}. These methods allocate pixels, attention, or detector calls within one scene, whereas we target the best retrieval score hidden below an unopened node.

\paragraph{MLLM zooming and visual agents.}
V*~\cite{wu2023vstar}, ZoomEye~\cite{shen2025zoomeye}, ZoomSearch~\cite{zhou2025zoomsearch}, and DeepEyes~\cite{zheng2025deepeyes} let multimodal large language models (MLLMs) acquire finer visual evidence during reasoning, and GeoVista extends this active perception to ultra-high-resolution remote-sensing scenes~\cite{zhu2026geovista} with dynamic-resolution backends such as Qwen2.5-VL~\cite{bai2025qwen25vl}. In each, an MLLM repeatedly picks which sub-region of one image to inspect next. Further earth-observation agents crop, zoom, or compress evidence within a single scene under a budget~\cite{liu2026zoomearth,wang2026geolens,wang2026geoeyes,dang2026uhrbat}, paying an MLLM forward for every glimpse of every query. We pursue retrieval rather than reasoning, and the bound decides where to look with a small head over embeddings the search has already computed, so per-query compute goes to encoding tiles rather than deciding among them.

\paragraph{Overhead retrieval and representation learning.}
EarthLoc localizes astronaut photographs by retrieval against a worldwide satellite gallery~\cite{berton2024earthloc}. Its reference gallery is fully pre-encoded, so EarthLoc reduces matching latency after index construction, whereas we study which gallery tiles need to be encoded for a query. Concurrently, Just Zoom In navigates overhead maps for cross-view geo-localization via autoregressive coarse-to-fine zooming~\cite{erzurumlu2026autoregressive}, but optimizes only for accuracy and with an unconstrained budget.

Remote-sensing coarse-to-fine retrieval operates over a fully pre-encoded index and can scale representation capacity~\cite{chen2026fastthenfine}. GeoDIVE makes encoding itself the budgeted resource and is optimized for that.

RemoteCLIP~\cite{liu2024remoteclip}, GeoRSCLIP~\cite{zhang2023rs5m}, and SkyCLIP~\cite{wang2024skyscript} adapt vision encoders to overhead imagery, DINO-style models provide strong instance features~\cite{oquab2023dinov2,simeoni2025dinov3}, and open-vocabulary detectors localize named objects within an opened image~\cite{minderer2023owlv2}. These are upstream representations rather than search policies.

\paragraph{Late interaction.}
ColBERT introduced token-level MaxSim for retrieval~\cite{khattab2020colbert}, and ColPali applies multi-vector late interaction to visually rich pages~\cite{faysse2024colpali}. We apply the same scoring principle only to patch tokens of tiles already visited by descent, so late interaction is a small online rerank rather than a patch-token index over the full gallery. Budget-elastic nested embeddings trade dot-product and storage cost at test time~\cite{xiang2026mmmatryoshka,xiao2025metaembed}. Our cost is the encoder forward itself, which nesting alone does not address.

\section{Benchmark and Protocol}
\label{sec:dataset}

\paragraph{Anchor-tile retrieval.}
An instance is a single annotated ground object, such as a building or a mapped area feature, and its \emph{anchor tile} is the highest resolution and deepest tile that fully contains it. A query is an instance crop from its target tile's imagery; captions of these crops provide a second, harder modality. The task is to return the object's anchor tile from the full gallery (Fig.~\ref{fig:descent-trace}). Objects span many spatial scales, so anchor tiles occur across zoom levels and the search must cover the whole pyramid. Each query has exactly one correct tile, which gives the bound of \S\ref{sec:method:bound} a single target. A caption keeps that unique answer but underspecifies it, so text queries probe the search under a weaker signal.

\paragraph{Data curation.}
Training imagery comes from Git-10M~\cite{liu2025text2earth}, a worldwide satellite corpus spanning zooms 10--18. We supplement it with over 100 million object instances generated from Microsoft building footprints and OpenStreetMap polygons, and the encoder adapters train only on this corpus. Git-10M's tiles are sampled across scales, so its hierarchies are sparse. A deployed archive instead serves complete pyramids, making a dense forest the more realistic search gallery.
We therefore construct the gallery from the retrieval split of Dense500~\cite{khanal2026genesis}, creating objects from 256 complete site pyramids containing over 20K tiles, and we ensure its imagery has no overlap with Git-10M. We filter to objects that cover at least 0.5\% of their anchor tile and test on 5{,}000 queries, with the remaining site trees serving as distractors. Finally, we caption a representative set of crops in each corpus, prompting the VLM with auxiliary information such as OSM tags, land cover, and settlement type.

\paragraph{Protocol and metrics.}
All experiments use three-fold cross-validation blocked on site, and each fold nests a validation partition inside its training pool, where every hyperparameter, including the stopping margin $\tau$, is selected and frozen. Queries are then evaluated over the same gallery. We report R@$k$ and MRR alongside search cost. Budget is calculated as a percent of gallery tiles encoded, acting as a proxy for the varied cost of node acquisition, encoding, and scoring. Each query keeps no embeddings from earlier queries, and root tiles count against its budget.

\section{Method}
\label{sec:methodology}
We outline the GeoDIVE framework for heuristic best-first search over satellite image forests. In the following sections, we describe the core components of GeoDIVE at both training and inference-time.

\subsection{A learned bound on descendant similarity}
\label{sec:method:bound}

Let $\mathbf{q}=f(Q)$ be the embedding of an object query $Q$ and $\mathbf{t}=f(T)$ the embedding of tile $T$. We denote the score function on a query-tile pair $(Q, T)$ using a cosine similarity as $\text{sim}(\mathbf{q},\mathbf{t})$. A query for $\mathbf{q}$ over a standard flat index on the set of all highest-resolution candidate tiles $\mathcal{T}$ requires computing $\argmax_{T \in \mathcal{T}} \text{sim}(\mathbf{q}, \mathbf{t})$, which incurs a large cost for large galleries that are required for large-scale, fine-grained search. We aim to mitigate this cost by discarding candidates in $\mathcal{T}$ that are estimated to be unimportant from their parent nodes. Our objective for learning an estimator useful for heuristic search follows from the observation that a tile $T$'s importance is defined by the most relevant tile in the subtree that $T$ defines. We denote this subtree as $\mathcal{C}(T)$, which is a set containing $T$ and all of its descendants. We then define the critical score 
\begin{equation}
c^*(\mathbf{q},T)=\max_{c\in\mathcal{C}(T)} \text{sim}(\mathbf{q},f(c)).
\label{eq:tstar}
\end{equation}
Note that this value can differ from the score $\text{sim}(\mathbf{q}, f(T))$ between tile $T$ and the query, as a coarse-grained tile can be a weak match for the query even if its descendants contain high resolution information that matches the query. 
Computing $c^*$ exactly requires encoding all descendants, which is computationally expensive, so we propose to estimate $c^*$ optimistically during the traversal process when observing an arbitrary tile. We estimate an optimistic value for $c^*$ with a scalar residual head on the baseline similarity,
\begin{equation}
\hat{c}(\mathbf{q},T)=\text{sim}(\mathbf{q},\mathbf{t})+
\operatorname{softplus}~ \!\bigl(h_\theta(\mathbf{q},\,\mathbf{t},\,\log |\mathcal{C}(T)|) \bigr),
\label{eq:bound}
\end{equation}
where $\operatorname{softplus} (u)=\log(1+e^{u})$ keeps the residual non-negative, $h_\theta$ is a two-layer MLP over the concatenated inputs. We denote the softplus term as $r = \operatorname{softplus}~ \!\bigl(h_\theta(\mathbf{q},\,\mathbf{t},\,\log |\mathcal{C}(T)|) \bigr)$. Note that $\log |\mathcal{C}(T)|$ provides a proxy for the size of the subtree under $T$, which allows the estimator to understand the amount of unobserved higher-frequency information. We strictly predict a positive residual anchored on the observed cosine similarity as confidence in a node cannot decrease after observing higher-resolution children of a node. When a leaf has no descendants, we set $\hat{c}(\mathbf{q},T)=\text{sim}(\mathbf{q},\mathbf{t})$ whenever $|\mathcal{C}(T)|=1$. We refer to the head without the $\log |\mathcal{C}(T)|$ input the \emph{scalar bound} and the full head the \emph{size-aware bound}.

\begin{figure*}[t]
\centering
\includegraphics[width=\textwidth]{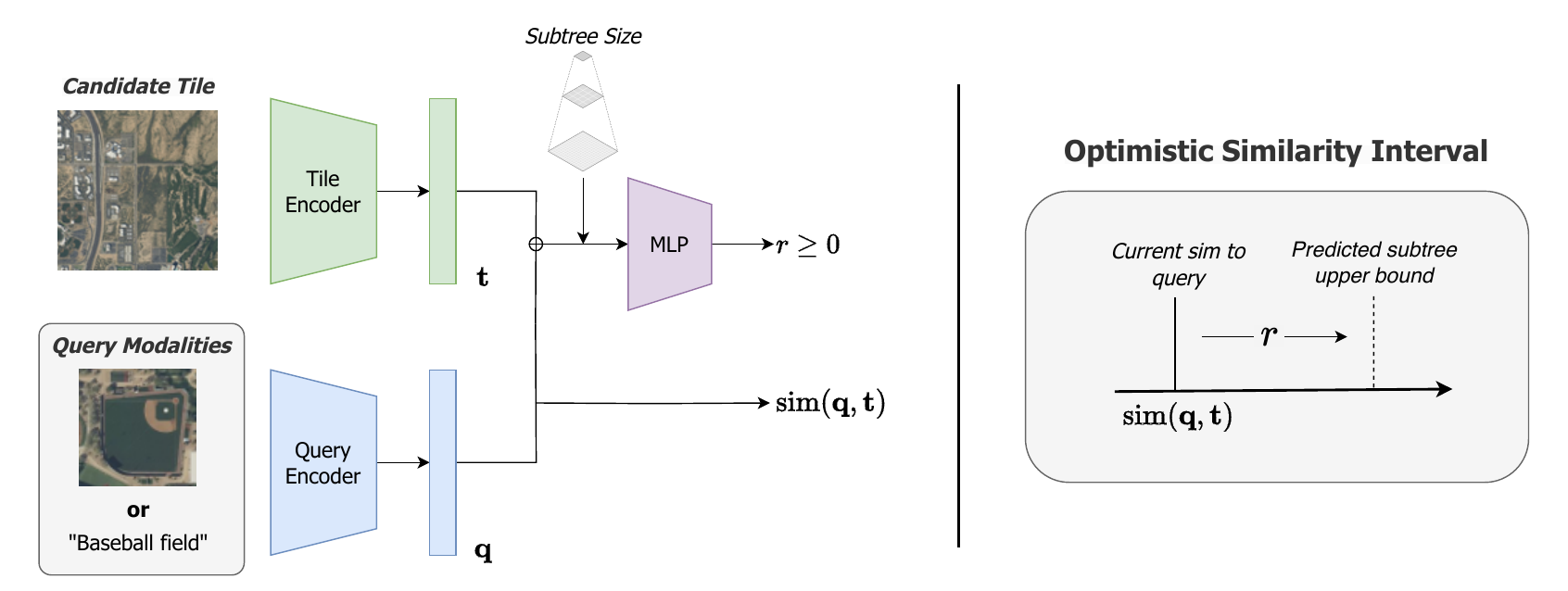}
\caption{\textbf{Bound architecture.} Image and text queries are evaluated separately: crop and tile use DINOv3, while caption and tile use GeoRSCLIP's respective towers. One path computes the cosine similarity $\text{sim}(\mathbf{q}, \mathbf{t})$; the other feeds both embeddings and subtree size to an MLP that outputs a residual $r$. Their sum is the traversal priority $\hat{c}(\mathbf{q}, \mathbf{t})$ of Eq.~\ref{eq:bound}, fit directly to the subtree-maximum target: the tile's own similarity plus how far above it the best hidden descendant can plausibly sit (right panel).}
\label{fig:architecture-bound}
\end{figure*}

\paragraph{An order-statistics prior.}
Under query $\mathbf{q}$, the descendants of a node $T$ carry $|\mathcal{C}(T)|$ scores $\{\text{sim}(\mathbf{q},f(c))\mid c\in\mathcal{C}(T)\}$. Assuming these scores obey an approximately Gaussian distribution, the expectation over the maximum target $c^*$ in Eq.~\ref{eq:tstar} is approximately:
\begin{equation}
    c^* \approx \mu+\kappa(n)\,\sigma,
\label{eq:blom}
\end{equation}
where $\kappa(n)$ is the expected largest of $n$ standard normal draws from distribution with mean $\mu$ and standard deviation $\sigma$. By assuming that subtree similarity scores are approximately Gaussian, we can adopt this form to parameterize the prediction of our optimistic gain. From this form, it intuitively follows that an increasing subtree size $n=|\mathcal{C}(T)|$ produces an increasingly optimistic gain, as a large subtree provides an increased area over which to observe relevant child imagery. As higher-resolution imagery cannot decrease confidence in a query-target match, we can interpret the learned residual term $\kappa(n)\,\sigma$ as how far the best hidden tile can plausibly sit above the visible score, given how uncertain the subtree's contents are and how many tiles it conceals. 
Discussion of Blom's closed-form $\kappa$ \cite{blom1958statistical} and the full calibration can be found in the supplement.

\subsection{Bound training}
\label{sec:method:training}

We train the model on pairs of an object query and a multi-resolution tile, each labeled with the regression target $c^*$ defined by Eq.~\ref{eq:tstar}. During training every target is computed from the cached embeddings, as the maximum over up to $M=24$ descendants sampled from the tile's subtree. The sample can underestimate the true maximum subtree similarity, approximately by the distribution discussed previously. For each query, the corresponding positive targets are the ancestors of its target tile. These tiles define the chain of correct tile selections a traversal should follow. For each positive pair we also sample two tiles uniformly from the training forest as negatives, as their subtrees do not contain the query target.

We regress the bound with a pinball loss,
\begin{equation}
\mathcal{L}_{\mathrm{bound}}
=\alpha\,[c^*-\hat{c}]_+ +(1-\alpha)\,[\hat{c}-c^*]_+,
\label{eq:pinball}
\end{equation}
where $[u]_+=\max(u,0)$. The asymmetry encodes the trade-off between accuracy and efficiency in the search. Underestimating $c^*$ can prune the branch that hides the answer, which costs recall. Overestimating it spends extra encoder calls, which costs efficiency. We set $\alpha=0.99$ so the loss fits the 99th conditional percentile of $c^*$, a near-upper estimate of the subtree maximum to approximate lossless search. Lowering $\alpha$ tightens the bound and raises the risk of pruning the answer, so $\alpha$ is the knob that trades the two \cite{futuhi2025admissible,sakaue2022sample}. Violation rates of the trained bound and comparisons with alternative training objectives are in the supplement.

\subsection{Budgeted traversal and adaptive stopping}
\label{sec:method:traversal}

For roots $\mathcal{R}$ and node budget $B$ (the encode fraction times the
gallery size), the descent maintains a frontier $\mathcal{F}$ of tiles
eligible for expansion:
\begin{algorithm}[t]
\caption{Budgeted best-first descent}
\label{alg:descent}
\begin{algorithmic}[1]
\State $V \gets \mathrm{encode}(\mathcal{R})$;\quad $\mathcal{F} \gets \mathcal{R}$ \Comment{roots seed the frontier}
\While{$|V| < B$ \textbf{and} $\mathcal{F} \neq \emptyset$}
  \State $T \gets \arg\max_{u\in\mathcal{F}} \hat{c}(\mathbf{q},u)$;\quad $\mathcal{F} \gets \mathcal{F}\setminus\{T\}$
  \For{each child $c$ of $T$ in the gallery}
    \State $\mathrm{encode}(c)$;\quad $V \gets V \cup \{c\}$ \Comment{one encode per child}
    \If{$c$ has children} $\mathcal{F} \gets \mathcal{F}\cup\{c\}$ \Comment{leaves never expand}
    \EndIf
  \EndFor
\EndWhile
\State \Return $V$ ranked by $\text{sim}(\mathbf{q},\cdot)$ \Comment{rank what was encoded}
\end{algorithmic}
\end{algorithm}
The traversal expands tiles in order of the bound $\hat{c}$, while the similarity $\text{sim}$ ranks the tiles already visited. An inflated bound can waste budget but cannot improve a tile's final rank. A tile that was never encoded cannot be recovered by any ranker, so reach, the fraction of queries whose target enters $V$, is decided entirely during traversal and caps every downstream metric. With $B$ equal to the gallery size every tile is encoded and the descent reduces to flat retrieval.

A fixed budget compares methods at equal cost. In deployment, an easy query should stop early and a hard one should search longer. We therefore also test an adaptive rule that replaces the budget test $|V| < B$ with a margin test, stopping the descent once no frontier bound can beat the best tile found so far,
\begin{equation*}
\max_{T\in\mathcal{F}}\hat{c}(\mathbf{q},T)\le \text{sim}_{\mathrm{best}}-\tau,
\end{equation*}
where $\text{sim}_{\mathrm{best}}$ is the best similarity among visited tiles. We select the cheapest margin $\tau$ whose validation reach is at least 95\%. The bound can underestimate $c^*$, so recall under adaptive stopping is measured rather than guaranteed. We report fixed-budget, adaptive, and oracle operation separately. The oracle stops with knowledge of the exact $c^*$, which is intractable in practice and serves purely as a reference. 

\subsection{Scoring visited tiles}
\label{sec:method:rerank}

Descent returns the visited set $V$, and scoring inside $V$ is standard retrieval. Cosine similarity ranks every visited tile. We optionally rerank the strongest candidates with MaxSim late interaction over patch tokens \cite{khattab2020colbert,faysse2024colpali}. The tokens are a byproduct of encoding the visited tiles, so reranking adds no encoder cost. A full late-interaction index would instead need patch tokens for the entire corpus. Late-interaction never changes $V$, and any gain is ranking only.

\subsection{Experimental details}
We evaluate \name{} on object image queries, following the protocol of \S\ref{sec:dataset}, and additionally on caption queries as a harder probe of the same search. Object queries use a satellite-pretrained DINOv3 ViT-L/16~\cite{simeoni2025dinov3}, and text queries use GeoRSCLIP ViT-L/14~\cite{zhang2023rs5m} with rank-16 LoRA~\cite{hu2022lora} on both towers, each adapted on Git-10M. Throughout, \name{} refers to the complete system, a best-first descent guided by the size-aware bound.

\subsection{Baselines}
We discuss baselines for \name{} based on heuristic search, trained policies for hierarchical traversal, and MLLM-driven agents. From our own system we report the scalar bound, which receives no subtree-size input, and raw-cosine descent, which predicts no bound at all. The ZoomEye schedule keeps ZoomEye's depth-blending formula~\cite{shen2025zoomeye} but drives it with the tile's own cosine as its ``is the target visible here'' confidence and the scalar bound as its ``is the target hidden below'' confidence. As trained search policies we use a behavior-cloned pointer that imitates oracle traversals~\cite{he2014learningtosearch}, OTM, which retrains its scorer on the states its own search visits~\cite{zhuo2020otm}, and VAS, a reinforcement-trained glimpse policy over the flat tile grid~\cite{sarkar2022vas}. Finally, we run the complete MLLM-driven ZoomEye system end to end (\S\ref{sec:results:comparisons}). 

\section{Results}
\label{sec:results}

\subsection{Quality at a budget}
\label{sec:results:main}

\begin{figure*}[t]
\centering
\includegraphics[width=\textwidth]{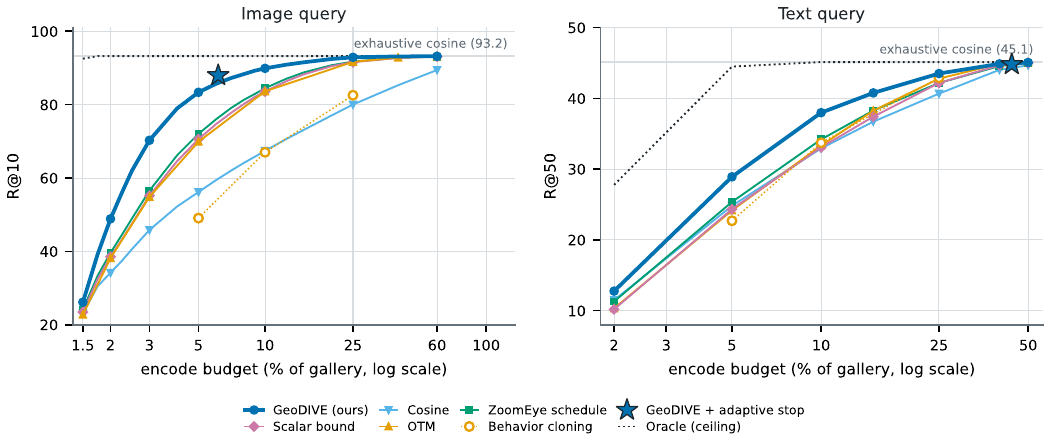}
\caption{\textbf{Encode--quality frontiers.} Image-crop queries at R@10
(left) and text-caption queries at R@50 (right) against encode budget. Every curve ranks its own visited set by a standard inner product. Stars mark the adaptive stop of \S\ref{sec:method:traversal}.}
\label{fig:pareto}
\label{fig:pareto-text}
\end{figure*}

Figure~\ref{fig:pareto} evaluates all methods across standardized encode budgets for both text and image query sets. Across all settings, \name is optimal in retrieval performance at matched encode budget, notably outperforming strong baselines based on reinforcement learning and MLLMs with significantly fewer accessed samples. While every curve across budgets asymptotically converges to the exhaustive ceiling, \name converges significantly faster and remains effective even at smaller budgets. For instance, on image queries \name with a 3\% encode budget approximately matches or exceeds all baselines at 5\% encode. Notably, at 25\% encoding, \name almost identically replicates the exhaustive cosine search performance, while requiring significantly less compute. Generally, we find that heuristic search policies produced by learning approximate search behaviors (OTM and behavior cloning) and MLLMs do not sufficiently capture a well-calibrated understanding of subtree value from identical pretrained embeddings.

Results across text queries reinforce this pattern (Fig.~\ref{fig:pareto-text}), showing that \name remains optimal across all encoding budgets, and quickly converges to near the exhaustive performance with only 25\% encoding. We note that in the text domain, we observe lower absolute recall metrics. This can be attributed to the inherent additional uncertainty in the limited text descriptions coming from OSM tag data, unlike the rich geometric precision given by an image query. Even so, the median target lands in the top 1\% of the full gallery, and we find that \name again outperforms all baselines across both RL and MLLM-based methods. 

\begin{figure*}[t]
\centering
\includegraphics[width=1\textwidth]{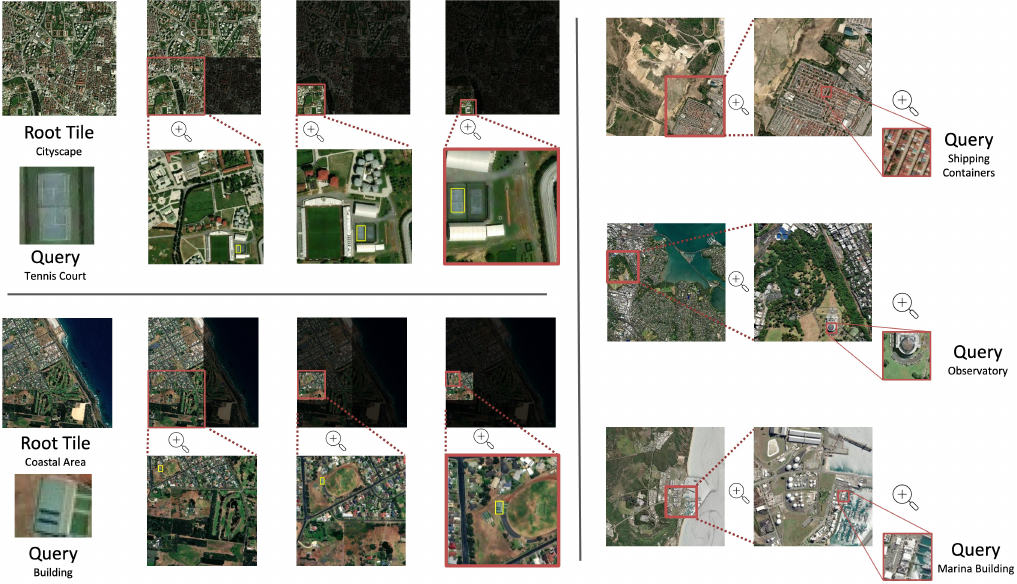}
\caption{\textbf{Qualitative traversals.} Left: two full descents. 
Darkened tiles were pruned without being encoded, red boxes mark the
followed branch, and yellow boxes mark candidate matches down to the
target. Right: three further queries, shown as retrieved site, descended
branch, and query.}
\label{fig:qualitative-traversal}
\end{figure*}

\subsection{Estimating the subtree maximum}
\label{sec:results:router}
To validate the design of \name's optimistic bound prediction, we study priorities derived from purely cosine, learned versus predetermined bounds, and the effect of adding structural signal (zoom level, subtree size) that the bound conditions on.

Each row of Table~\ref{tab:router} replaces the inputs to subtree maximum prediction. Variance composes the law of Eq.~\ref{eq:blom} directly, estimating a mean and spread for each subtree and adding $\kappa(n)\,\sigma$. Node zoom gives the learned head the tile's absolute zoom in place of subtree size, and the ZoomEye schedule weights node depth and the calculated residual, mirroring how the original ZoomEye blends its MLLM's answers to the same two questions~\cite{shen2025zoomeye}. Every row runs the same best-first search at 10\% encode with the same encoder.

The two query modalities are consistent across ablations. Every structured estimate improves on raw cosine, and the scalar bound and the depth-based schedules land within a point of one another in both modalities. Adding subtree size separates from all of them, gaining 6.3 R@10 over the scalar bound on object queries and 4.9 R@50 on text. %

\begin{table}[t]
\centering
\caption{\textbf{Estimator ablation.} Each row replaces the estimate of the subtree maximum that orders expansion. The search, the folds, the encoders, and the cosine ranking of visited tiles are fixed at 10\% encode. Rows prefixed $+$ each add one signal to
the scalar bound and do not stack. The oracle substitutes the exact subtree maximum for the estimate.}
\label{tab:router}
\footnotesize
\setlength{\tabcolsep}{3pt}
\begin{tabular}{@{}lrrr|rrr@{}}
\toprule
& \multicolumn{3}{c}{Object image $\to$ tile} & \multicolumn{3}{c}{Object text $\to$ tile} \\
\cmidrule(lr){2-4} \cmidrule(l){5-7}
Estimate of $c^*$ & R@5 & R@10 & MRR & R@50 & R@100 & MRR \\
\midrule
Cosine & 62.5 & 67.4 & 52.8 & 32.9 & 41.7 & 8.0 \\
Variance & 75.2 & 81.0 & 63.4 & 32.3 & 40.9 & 8.2 \\
Node zoom & 78.3 & 84.3 & 66.0 & 33.1 & 41.7 & 8.3 \\
Scalar bound & 77.6 & 83.6 & 65.4 & 33.1 & 41.7 & 8.4 \\
$+$ ZoomEye sched.~\cite{shen2025zoomeye} & 78.4 & 84.5 & 66.1 & 33.6 & 42.7 & 8.6 \\
$\mathbf{+}$ \textbf{subtree size} $\mathbf{n}$ \textbf{(ours)} & \textbf{83.4} & \textbf{89.9} & \textbf{70.3} & \textbf{38.0} & \textbf{48.2} & \textbf{9.3} \\
\midrule
\emph{Oracle} (exact subtree) & \emph{86.5} & \emph{93.2} & \emph{72.9} & \emph{45.1} & \emph{60.1} & \emph{10.8} \\
\bottomrule
\end{tabular}
\end{table}

The two learned heads differ in the residual they predict. Each head adds a residual above the node's own cosine, its estimate of how far the best hidden tile sits above the visible one. The true gap widens with the number of hidden tiles (Eq.~\ref{eq:blom}), and only the size-aware head reproduces that widening. Its residuals grow with subtree size at the rate of the 99th-percentile target, which the supplement plots for both heads. The scalar bound never sees $n$, so its residual stays nearly flat and it underestimates exactly the subtrees that hide the most. In effect, the small MLP has learned how far the best of $n$ hidden tiles should sit above the visible score.

\subsection{Reranking the visited set}
\label{sec:results:rerank}

\begin{figure*}[t]
\centering
\includegraphics[width=\textwidth]{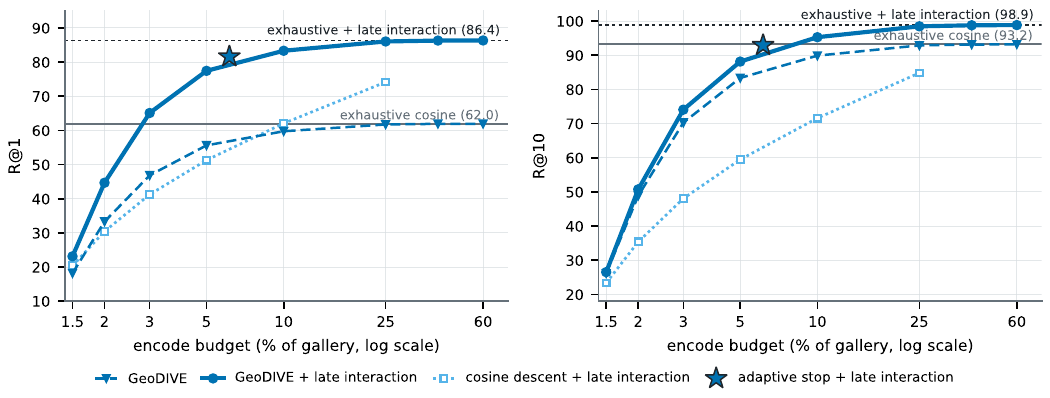}
\caption{\textbf{Results on reranking.} R@1 (left) and R@10 (right)
by encode budget for \name with and without reranking. Star: the adaptive stop of \S\ref{sec:method:traversal}.}
\label{fig:rerank-frontier}
\end{figure*}

For object image queries, we find that late interaction-based reranking provides significant performance gains without additional overhead cost. By reranking the 200-strongest candidates determined by the initial inner product, multi-vector late interaction reuses already-encoded tile tokens to improve performance (Table~\ref{tab:rerank}, Fig.~\ref{fig:rerank-frontier}). This free reranking allows the \name to achieve significant performance gains against baselines at a tenth of the encode, relative to mildly worse performance without second-stage reranking. Relative to the exhaustive and intractable MaxSim baseline, both our 10\% encode and adaptive stopping variants of \name achieve a significant share of the gain over inner product-scored search. This motivates our efficient tree-driven design, which enables this performant reranking without significant deployment overhead.

Figure~\ref{fig:rerank-frontier} traces this effect across the full budget range. Ranked by cosine alone, \name{} can at best approach the exhaustive-cosine ceiling, but with late interaction over the same visited tiles it passes that ceiling at 3\% encode on R@1 and 10\% on R@10, closing on the full-gallery late-interaction ceiling by 25\% without ever building a patch-token index. Applying the same reranker to cosine descent shows the gain is not the reranker alone: with late interaction on both sides, \name{} still leads by more than 23 R@10 at 10\% encode, so the visited set, not the scorer, decides the outcome. The adaptive stop composes both, landing within five R@1 points of the full-gallery ceiling at 6\% mean encode.

\begin{table}[t]
\centering
\caption{\textbf{Late-interaction reranking results.} Each
$+$ MaxSim row rescores the cosine top 200 of the visited set above it with a ColBERT-style MaxSim and
adds no additional encoder inference. Adaptive encode is the mean over queries. The
exhaustive $+$ MaxSim row requires patch tokens for the entire gallery.}
\label{tab:rerank}
\small
\setlength{\tabcolsep}{4pt}
\begin{tabular}{@{}lrrrr@{}}
\toprule
Method & Encode & R@1 & R@5 & R@10 \\
\midrule
\name{} & 10\% & 59.8 & 83.4 & 89.9 \\
$+$ MaxSim & & \textbf{83.3} & \textbf{93.8} & \textbf{95.3} \\
\addlinespace
\name{} + Adaptive Stop & 6.1\% & 59.0 & 81.9 & 88.0 \\
$+$ MaxSim & & \textbf{81.6} & \textbf{91.6} & \textbf{93.0} \\
\addlinespace
\hline
\addlinespace
Exhaustive cosine & 100\% & 62.0 & 86.5 & 93.2 \\
$+$ MaxSim & & \textbf{86.4} & \textbf{97.3} & \textbf{98.9} \\
\bottomrule
\end{tabular}
\end{table}

\begin{table}[t]
\centering
\caption{\textbf{R@10 at 5\% encode by object size} (fraction of tile
area). Baseline adaptations are detailed in the supplement.}
\label{tab:pending-comparisons}
\footnotesize
\setlength{\tabcolsep}{4pt}
\begin{tabular*}{\linewidth}{@{\extracolsep{\fill}}lrrrr|r@{}}
\toprule
System & $<$1\% & 1--2\% & 2--5\% & $\geq$5\% & all \\
Unique queries & 2{,}582 & 1{,}311 & 783 & 324 & 5{,}000 \\
\midrule
VAS~\cite{sarkar2022vas} & 5.8 & 5.8 & 6.7 & 6.2 & 6.0 \\
Cosine & 57.0 & 52.9 & 55.2 & 65.1 & 56.2 \\
Behavior Cloning~\cite{he2014learningtosearch} & 50.6 & 47.4 & 47.1 & 61.5 & 49.9 \\
OTM~\cite{zhuo2020otm} & 70.0 & 68.1 & 68.6 & 79.1 & 69.9 \\
Scalar bound & 70.7 & 68.9 & 69.4 & 79.6 & 70.6 \\
ZoomEye sched.~\cite{shen2025zoomeye} & 71.4 & 71.0 & 71.8 & 81.4 & 72.0 \\
\midrule
\textbf{\name{}} & \textbf{79.3} & \textbf{86.6} & \textbf{88.4} & \textbf{90.3} & \textbf{83.4} \\
\midrule
\emph{Oracle} & \emph{90.3} & \emph{94.4} & \emph{98.1} & \emph{99.7} & \emph{93.2} \\
\bottomrule
\end{tabular*}
\end{table}

\subsection{Comparisons to learned search policies}
\label{sec:results:comparisons}
Table~\ref{tab:pending-comparisons} compares \name{} with learned search policies under the full protocol, stratified by the object's fraction of its tile. We see across all query size strata that \name achieves the strongest performance. Notably, \name{} leads every alternative by more than 11 R@10 at 5\% encode, and its largest margins fall in the 1--2\% and 2--5\% object sizes.

We also run the complete ZoomEye system~\cite{shen2025zoomeye} with two MLLM backbones LLaVA-OV-7B and Qwen3-VL-8B. We adapt their pipeline for agentic hierarchical traversal to select the frontier candidates and encoding actions for search (see supplement). Most notably, the MLLM backbones are extremely computationally expensive and infeasible for realistic search.

Both models traverse over the correct answer tile (56\% and 64\% of queries at a 25\% budget), but the confidence score output by the MLLM results in less than 4\% R@10 over any budget. Using cosine similarity as the ranking metric for answer tiles fixes this problem and results in recall that equals the percent of the time the answer node is traversed. Fixing this does not change the deployment and inference demands -- LLaVA required at least 9{,}584 model forwards per query, and a single query takes a median of 2.8 hours end to end on one H100, against 1.72s for \name{} at the 5\% budget.

\section{Conclusion}
\label{sec:conclusion}

We introduced \name{}, an approach for geospatial search over large-scale imagery of queries ranging from a single building to a whole airport. \hl{In scenarios where exhaustive dense retrieval is intractable, \name{} recasts retrieval as traversal, descending the multi-resolution pyramid best-first and encoding a tile only when the search visits it.} This bypasses the enormous overhead of encoding massive galleries and avoids storing the dense patch-level vectors that richer scoring would otherwise require. A learned probabilistic estimate of the value of descending below each coarse tile combines what the tile shows with how much imagery its subtree conceals, so the search descends fully on matching branches and stops early on the rest.

We evaluated \name on densely annotated hierarchical imagery with object-level instances and captions, and show that \name{} reaches nearly exhaustive accuracy at a fraction of the encoding. \name's efficient reranking design provides further improvements to performance without significant overhead cost.

We leave a number of technical directions open to future work. These include further study for other hierarchical data domains (e.g. gigapixel histological data, small object search, aerial video, etc.), development of more sophisticated adaptive termination criteria, and multi-target search.

\clearpage
\section*{Acknowledgments}
This research used the TGI RAILs advanced compute and data resource which is supported by the National Science Foundation (award OAC-2232860) and the Taylor Geospatial Institute.

This material is based upon work supported in whole or in part with funding from the Department of Defense (DoD). Any opinions, findings, conclusions, or recommendations expressed in this material are those of the author(s) and do not necessarily reflect the views of the DoD and/or any agency or entity of the United States Government.

{
    \small
    \bibliographystyle{ieeenat_fullname}
    \bibliography{main}
}
\clearpage
\appendix
\section*{Supplementary Material}

\section{Data}
\label{supp:data}

\begin{figure*}[tbp]
\centering
\includegraphics[width=0.93\textwidth]{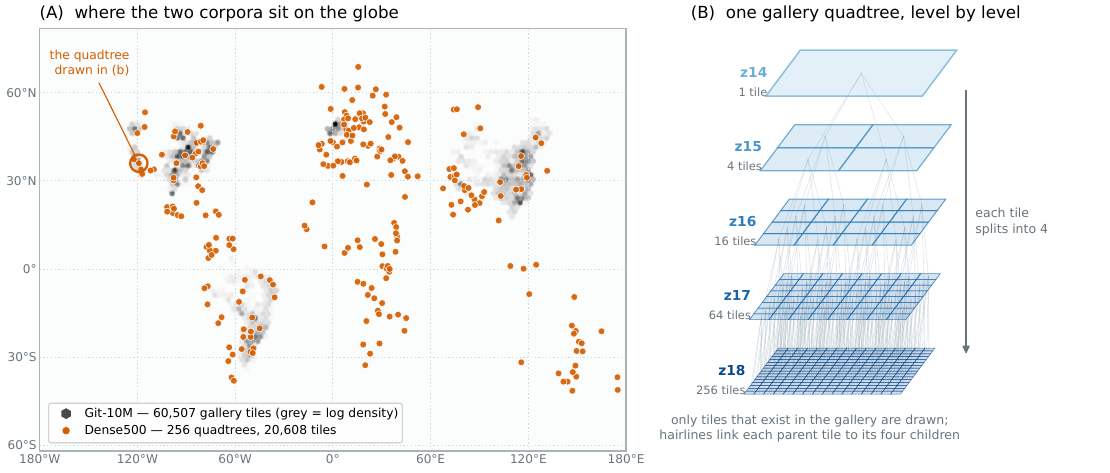}
\caption{\textbf{The gallery is a forest of spatial quadtrees.} 20{,}608 tiles in 256 disjoint trees; a coarse tile's four children partition its footprint at the next zoom.}
\label{fig:corpus-map}
\end{figure*}

A query is one object crop or its open-vocabulary text description, and
its unique target is the anchor tile in the Dense500 gallery. Except where
noted, tables report the object-crop modality. The gallery is a forest of
20{,}608 tiles in 256 site trees, with roots at zooms 10--15 and depths of
two to four edges (three to five zoom levels, including the root).
Figure~\ref{fig:corpus-map} shows the geographic
distribution. The gallery contains every tile of every site and is never
split. The 5{,}000 queries are drawn with a fixed seed from the 41{,}646
objects whose polygon covers at least 300\,px$^2$ on the 256-px tile
frame. Object annotations come from Microsoft building footprints and
OpenStreetMap polygons, with polygons above 20{,}000\,m$^2$ emitted as
area instances. Crops pad the object box by 0.15 of its long side, with a
4-px minimum, and expand to a square.

\paragraph{Pixel resolution and query scale.}
The native gallery tiles are $256\times256$ pixels. An image query is a
variable-size square crop determined by its object box and the padding
rule above, rather than a fixed-size patch from the source image. For the
main image-query encoder, both crops and gallery tiles are resized to
$224\times224$ pixels before encoding. These pixel dimensions are
distinct from ground sampling distance: tiles at different zooms cover
different geographic extents. The benchmark supplies no target zoom or
ground sampling distance to the retrieval system, and searches all
available levels. Reliable target-scale metadata could instead constrain
the candidate levels; performance under such a constraint is not
evaluated here.

\paragraph{Scope of the image-query benchmark.}
Image queries are crops from the same source imagery as their gallery
targets. The task therefore measures within-source instance retrieval and
the encoding cost of finding the corresponding anchor tile. Shared image
content can make matching easier than retrieval from an independently
acquired observation of the same object. The reported results do not
establish robustness to changes in acquisition date, illumination, cloud
cover, viewing angle, or sensor. Caption queries change the query modality
but are still derived from these crops. Likewise, the synthetic query
perturbations described below do not substitute for a cross-date or
cross-sensor evaluation.

\paragraph{Dense500 retrieval.}
Dense500~\cite{khanal2026genesis} releases 500 fully observed depth-4
quadtrees for multi-scale tile completion. Our gallery is its retrieval
split: 256 sites drawn with the same population-stratified,
continent-balanced, Git-10M-disjoint site sampler, but with site trees of
three to five zoom levels (roots at zooms 10--15). Variable
depth is used to simulate realistic conditions. Additionally, in a fixed-depth forest, subtree size is a function of
zoom alone and the subtree-size term of the main paper's bound could not be
separated from a depth schedule. 

\section{Text-query track}
\label{supp:text-track}

\begin{figure*}[t]
\centering
\includegraphics[width=\textwidth]{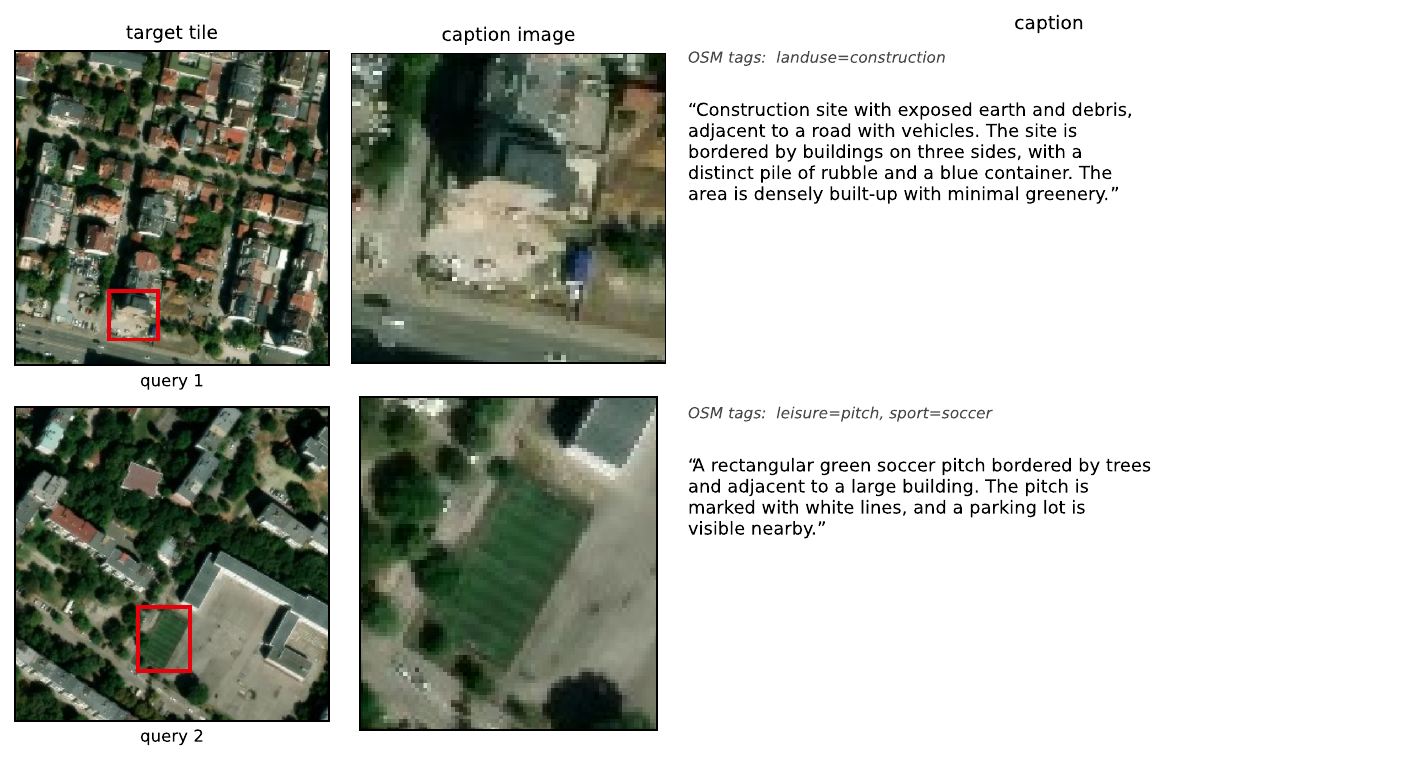}
\caption{\textbf{Caption queries.} Left: the target tile with the instance
box marked. Centre: the padded crop shown to the captioner. Right: the OSM
tags supplied as context and the resulting caption, which is the query.
Tags never appear in the query. The prompt asks for details visible in the
crop, such as the rubble pile and blue container (top) and the white pitch
lines (bottom).}
\label{fig:caption-example}
\end{figure*}

Text queries are VLM captions of the object crops
(Fig.~\ref{fig:caption-example}). InternVL3-38B,
decoding greedily, is prompted with the crop plus the instance's OSM tags,
a land-cover summary, and a settlement class, so captions can name
functions that are not visible in the pixels. The tags themselves are
never the query. Caption crops use the same 0.15 padding as image queries
with a larger 16-px minimum. The text track uses GeoRSCLIP
ViT-L/14 with rank-16 LoRA on both towers, adapted on Git-10M.

\paragraph{Separate query modalities.}
There is no fusion of an image query and a text query into a single
embedding. In the image track, the crop and gallery tile pass through the
same adapted DINOv3 image encoder. In the text track, the caption passes
through the GeoRSCLIP text tower and the gallery tile through its image
tower. Cosine similarity is computed within the corresponding embedding
space, and the bound head receives that track's query and tile embeddings
together with subtree size. DINOv3 image-query embeddings are not compared
with GeoRSCLIP gallery embeddings. The crop and auxiliary metadata are
inputs to caption generation; the resulting caption alone is the text
retrieval query.

The evaluation uses 4,000 caption queries over three seeds of the
site-blocked folds. Text metrics lead with R@50 because captions
underspecify their targets. The exhaustive-cosine ceiling is 45.13 R@50.
Table~\ref{tab:text-grid} gives the budget grid for the deployable
priorities beside the oracle.

\begin{table}[htbp]
\centering
\caption{\textbf{Text-query budget grid.} Reach and R@50 by encode budget.}
\label{tab:text-grid}
\scriptsize
\setlength{\tabcolsep}{3pt}
\begin{tabular}{@{}rrrrrrrrr@{}}
\toprule
& \multicolumn{2}{c}{Cosine} & \multicolumn{2}{c}{Scalar bound}
& \multicolumn{2}{c}{Size-aware} & \multicolumn{2}{c}{\emph{Oracle}} \\
Enc.\ \% & Reach & R@50 & Reach & R@50 & Reach & R@50 & \emph{Reach} & \emph{R@50} \\
\midrule
2 & 11.63 & 11.48 & 10.28 & 10.19 & 12.81 & 12.76 & \emph{28.17} & \emph{27.78} \\
5 & 33.44 & 24.81 & 31.14 & 24.34 & 38.15 & 28.91 & \emph{62.03} & \emph{44.49} \\
10 & 54.89 & 32.93 & 52.90 & 33.08 & 62.34 & 37.98 & \emph{82.53} & \emph{45.13} \\
15 & 67.62 & 36.71 & 67.23 & 37.41 & 74.91 & 40.77 & \emph{90.20} & \emph{45.13} \\
25 & 81.67 & 40.67 & 84.22 & 42.18 & 88.54 & 43.51 & \emph{96.55} & \emph{45.13} \\
40 & 92.84 & 44.03 & 94.80 & 44.68 & 96.48 & 44.88 & \emph{99.23} & \emph{45.13} \\
50 & 96.06 & 44.67 & 97.50 & 45.02 & 98.56 & 45.07 & \emph{99.58} & \emph{45.13} \\
\bottomrule
\end{tabular}
\end{table}

Table~\ref{tab:text-towers} swaps the text tower with the protocol fixed. Rank-16 LoRA on
GeoRSCLIP lifts the ceiling to 45.13, so adaptation matters more than
domain pretraining. Size-aware descent beats raw-cosine descent at 10\%
encode on five of the six towers.

\begin{table}[htbp]
\centering
\caption{\textbf{Text-tower swap.} R@50 on the 74-site text setting with the tile tower and protocol fixed: the exhaustive ceiling and the descent priorities at 10\% encode.}
\label{tab:text-towers}
\scriptsize
\setlength{\tabcolsep}{2pt}
\begin{tabular}{@{}lrrrrr@{}}
\toprule
& & \multicolumn{4}{c}{R@50 at 10\% encode} \\
\cmidrule(lr){3-6}
Text tower & Exhaustive & Cosine & Scalar & Size-aware & $\Delta$ \\
\midrule
CLIP ViT-L/14 & 11.92 & 9.64 & 9.64 & 9.71 & +0.07 \\
OpenCLIP ViT-L/14 & 15.24 & 12.72 & 12.54 & 14.01 & +1.29 \\
RemoteCLIP ViT-L/14 & 14.35 & 12.56 & 11.78 & 12.48 & -0.08 \\
GeoRSCLIP ViT-L/14 & 15.92 & 13.22 & 13.95 & 13.96 & +0.74 \\
SigLIP2 so400m & 15.17 & 12.52 & 13.64 & 13.78 & +1.26 \\
\midrule
\textbf{GeoRSCLIP $+$ LoRA} & \textbf{45.13} & \textbf{32.93} & \textbf{33.08} & \textbf{37.98} & +5.05 \\
\bottomrule
\end{tabular}
\end{table}

\section{Extended object-query results}
\label{supp:v3-tables}

\begin{table*}[tbp]\centering
\caption{\textbf{The size input across encoders.} R@10 at 10\% encode with only the encoder changed; \emph{Closed} is the fraction of the scalar-to-oracle gap that the size input recovers.}
\label{tab:encoders}
\small
\begin{tabular}{@{}lrrrrrr@{}}
\toprule
Encoder & Cosine & Scalar & Size-aware & Oracle & Gain & Closed \\
\midrule
\textbf{DINOv3-SAT $+$ our LoRA} & \textbf{67.4} & \textbf{83.6} & \textbf{89.9} & \textbf{93.2} & \textbf{+6.3} & \textbf{66\%} \\
DINOv3-SAT, frozen & 30.0 & 31.9 & 33.9 & 36.7 & +1.97 & 42\% \\
GeoRSCLIP ViT-L/14~\cite{zhang2023rs5m} & 7.3 & 8.3 & 8.6 & 9.5 & +0.36 & 30\% \\
RemoteCLIP ViT-L/14~\cite{liu2024remoteclip} & 4.6 & 5.6 & 6.2 & 8.8 & +0.62 & 19\% \\
DINOv2 ViT-L/14 & 11.9 & 19.2 & 21.6 & 25.0 & +2.43 & 42\% \\
CLIP ViT-B/32 & 5.7 & 6.4 & 6.6 & 7.6 & +0.23 & 20\% \\
\bottomrule
\end{tabular}
\end{table*}

\begin{table*}[tbp]
\centering
\caption{\textbf{Size-aware minus scalar bound by object area and by target depth.} Paired deltas with site-blocked 95\% CIs.}
\label{tab:strata}
\scriptsize
\setlength{\tabcolsep}{4pt}
\begin{tabular}{@{}llrrrr@{}}
\toprule
Enc. & Area (px$^2$) & $N$ & $\Delta$ Reach & $\Delta$ R@1 & $\Delta$ R@10 \\
\midrule
5\% & [300, 600) & 6,990 & +9.2 {\scriptsize [5.2, 13.0]} & +3.9 {\scriptsize [2.1, 5.8]} & +7.7 {\scriptsize [4.3, 11.2]} \\
 & [600, 1200) & 4,266 & +17.3 {\scriptsize [11.0, 22.6]} & +10.2 {\scriptsize [7.0, 13.4]} & +15.8 {\scriptsize [10.2, 20.8]} \\
 & [1200, 2400) & 2,256 & +20.7 {\scriptsize [12.9, 27.3]} & +15.2 {\scriptsize [9.6, 19.8]} & +19.7 {\scriptsize [12.4, 26.0]} \\
 & [2400, inf) & 1,488 & +11.7 {\scriptsize [5.0, 17.4]} & +10.7 {\scriptsize [4.5, 16.1]} & +11.7 {\scriptsize [5.0, 17.4]} \\
\midrule
10\% & [300, 600) & 6,990 & +4.7 {\scriptsize [2.3, 7.5]} & +2.3 {\scriptsize [0.9, 4.1]} & +4.2 {\scriptsize [2.0, 7.0]} \\
 & [600, 1200) & 4,266 & +7.9 {\scriptsize [3.8, 12.2]} & +4.2 {\scriptsize [2.0, 6.7]} & +6.8 {\scriptsize [3.1, 10.9]} \\
 & [1200, 2400) & 2,256 & +8.7 {\scriptsize [4.8, 13.5]} & +6.7 {\scriptsize [3.6, 10.2]} & +8.3 {\scriptsize [4.6, 13.0]} \\
 & [2400, inf) & 1,488 & +8.1 {\scriptsize [3.5, 13.4]} & +7.2 {\scriptsize [3.0, 12.1]} & +8.1 {\scriptsize [3.5, 13.4]} \\
\midrule
25\% & [300, 600) & 6,990 & +1.5 {\scriptsize [0.5, 2.7]} & +0.7 {\scriptsize [0.2, 1.5]} & +1.3 {\scriptsize [0.4, 2.4]} \\
 & [600, 1200) & 4,266 & +1.0 {\scriptsize [0.4, 1.8]} & +0.6 {\scriptsize [0.1, 1.2]} & +0.9 {\scriptsize [0.3, 1.7]} \\
 & [1200, 2400) & 2,256 & +0.9 {\scriptsize [0.3, 2.0]} & +0.8 {\scriptsize [0.3, 1.5]} & +0.9 {\scriptsize [0.3, 2.0]} \\
 & [2400, inf) & 1,488 & +1.6 {\scriptsize [0.4, 3.4]} & +1.5 {\scriptsize [0.3, 3.1]} & +1.6 {\scriptsize [0.4, 3.4]} \\
\bottomrule
\end{tabular}\hfill
\begin{tabular}{@{}llrr@{}}
\toprule
Enc. & Depth & $N$ & $\Delta$ R@10 \\
\midrule
5\% & Depth 0 & 249 & +0.0 {\scriptsize [0.0, 0.0]} \\
 & Depth 1 & 486 & +1.6 {\scriptsize [0.0, 4.8]} \\
 & Depth 2 & 1,590 & +0.2 {\scriptsize [-3.8, 4.4]} \\
 & Depth 3 & 8,271 & +7.5 {\scriptsize [2.6, 13.1]} \\
 & Depth 4 & 4,404 & +27.2 {\scriptsize [19.2, 37.9]} \\
 & Leaf targets & 12,534 & +13.6 {\scriptsize [8.1, 18.7]} \\
\midrule
10\% & Depth 0 & 249 & +0.0 {\scriptsize [0.0, 0.0]} \\
 & Depth 1 & 486 & +0.8 {\scriptsize [0.0, 2.0]} \\
 & Depth 2 & 1,590 & +2.1 {\scriptsize [0.0, 5.8]} \\
 & Depth 3 & 8,271 & +4.6 {\scriptsize [1.6, 8.2]} \\
 & Depth 4 & 4,404 & +10.9 {\scriptsize [4.8, 20.8]} \\
 & Leaf targets & 12,534 & +6.5 {\scriptsize [3.5, 10.2]} \\
\midrule
25\% & Depth 0 & 249 & +0.0 {\scriptsize [0.0, 0.0]} \\
 & Depth 1 & 486 & +0.6 {\scriptsize [0.0, 2.1]} \\
 & Depth 2 & 1,590 & +0.5 {\scriptsize [0.0, 1.1]} \\
 & Depth 3 & 8,271 & +1.5 {\scriptsize [0.5, 2.8]} \\
 & Depth 4 & 4,404 & +0.9 {\scriptsize [0.2, 2.3]} \\
 & Leaf targets & 12,534 & +1.3 {\scriptsize [0.6, 2.2]} \\
\bottomrule
\end{tabular}
\end{table*}

Table~\ref{tab:encoders} tests whether the subtree-size input helps only
our adapted encoder or any encoder. Each row swaps the encoder and keeps
the corpus, tree and queries fixed. The size-aware bound beats the scalar
bound for all six encoders, closing 19--66\% of each one's gap to the
oracle.

Table~\ref{tab:strata} tests where the size input helps most. It splits
the size-aware minus scalar gain by object area and by target depth. The
gain is largest for deep targets at tight budgets ($+27.2$ R@10 at depth 4
and 5\% encode) and fades as the budget opens more of the tree.

Table~\ref{tab:main-full} tests every system at its own encode cost on the
same gallery and queries, with R@1/5/10 and MRR (equal to mAP with one
target per query). It shows \name{} at 5\% encode within 1.2 R@10 of the
strongest 10\%-budget alternative, the ZoomEye schedule, and above every
other one, and \name{} at 10\% above all of them by at least 5 R@10. The
adaptive stop (6.14\% mean encode, 93.91\% reach) lands within 0.3 R@10 of
exhaustive cosine once reranked. The oracle
row runs the same stop with the exact subtree maximum and is not
deployable.

\begin{table*}[tbp]
\centering
\caption{\textbf{Extended object-to-tile quality at budget.} Encoder work priced at batch-64 A100 throughput (2.179\,ms per forward). The cascade row is priced in forward-equivalents; adaptive rows report achieved mean encode.}
\label{tab:main-full}
\small
\setlength{\tabcolsep}{4pt}
\begin{tabular}{@{}lrrrrrrr@{}}
\toprule
System & Encode & Time (s) & TFLOPs & R@1 & R@5 & R@10 & MRR \\
\midrule
Uniform subsample & 10.0\% & 4.5 & 259 & 8.6 & 9.7 & 9.8 & .091 \\
Single coarse level & 5.0\% & 2.2 & 129 & 2.9 & 3.2 & 3.2 & .031 \\
TDM tile classifier~\cite{zhu2018tdm} & 10.0\% & 4.5 & 259 & 40.5 & 56.5 & 60.7 & .476 \\
Raw cosine descent & 10.0\% & 4.5 & 259 & 44.8 & 62.5 & 67.4 & .528 \\
Cascade ($K{=}100$) & 21.3\% & 9.6 & 551 & 53.2 & 72.5 & 77.0 & .617 \\
\textbf{\name} (5\% budget) & \textbf{5.0\%} & \textbf{2.2} & \textbf{130} & \textbf{55.63} & \textbf{77.51} & \textbf{83.37} & \textbf{.654} \\
\quad $+$ late interaction & & & & \textbf{77.45} & \textbf{86.95} & \textbf{88.16} & \textbf{.816} \\
ZoomEye schedule~\cite{shen2025zoomeye} & 10.0\% & 4.5 & 259 & 56.2 & 78.4 & 84.5 & .661 \\
\textbf{\name} $+$ adaptive stop & \textbf{6.14\%} & \textbf{2.76} & --- & \textbf{59.03} & \textbf{81.89} & \textbf{87.98} & \textbf{.692} \\
\emph{Oracle} $+$ adaptive stop & \emph{1.47\%} & \emph{0.66} & --- & \emph{61.96} & \emph{86.13} & \emph{92.51} & \emph{.727} \\
\textbf{\name} (10\% budget) & \textbf{10.0\%} & \textbf{4.5} & \textbf{259} & \textbf{59.76} & \textbf{83.43} & \textbf{89.90} & \textbf{.703} \\
\quad $+$ late interaction & & & & \textbf{83.33} & \textbf{93.83} & \textbf{95.30} & \textbf{.880} \\
\textbf{\name} (25\% budget) & \textbf{25.0\%} & \textbf{11.2} & \textbf{648} & \textbf{61.75} & \textbf{86.23} & \textbf{92.90} & \textbf{.727} \\
\quad $+$ late interaction & & & & \textbf{86.05} & \textbf{96.93} & \textbf{98.48} & \textbf{.909} \\
\midrule
\emph{Exhaustive cosine} & \emph{100\%} & \emph{44.9} & \emph{2590} & \emph{62.0} & \emph{86.5} & \emph{93.2} & \emph{.729} \\
\bottomrule
\end{tabular}
\end{table*}

The $+$ late interaction rows of Table~\ref{tab:main-full} rerank only the
tiles the descent already encoded (MaxSim over the cosine top 200 within
the visited set) and add no encoder work. They show gains of 22--24 R@1 and
5--6 R@10 at every budget, all paired intervals excluding zero. From 10\%
encode the reranked descent exceeds exhaustive cosine at R@1 (by 21.4) and
matches it at R@10.

\section{Calibration and evaluation protocol}
\label{supp:statistical-corpus}

\subsection{Subtree-size calibration}

\begin{figure}[htbp]
\centering
\includegraphics[width=\linewidth]{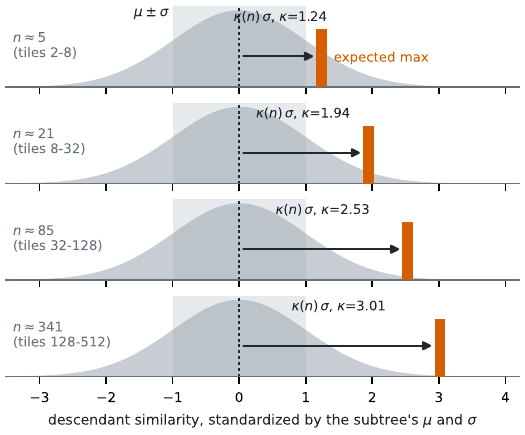}
\caption{\textbf{The subtree maximum rises with subtree size.} Descendant similarities standardized per subtree; the expected maximum of $n$ draws sits $\kappa(n)\,\sigma$ above the mean, with $\kappa$ measured per size bucket (Table~\ref{tab:blom-calibration}). Densities are illustrative.}
\label{fig:order-stats}
\end{figure}

\begin{table}[htbp]
\centering
\caption{\textbf{Order-statistic calibration on 750{,}000 cached query--tile pairs.} $\kappa$ is the standardized gap to the subtree maximum; the last two columns are the mean and 99th percentile of $c^*{-}\text{sim}$.}
\label{tab:blom-calibration}
\scriptsize
\setlength{\tabcolsep}{4pt}
\begin{tabular}{@{}lrrrrr@{}}
\toprule
$n$ bucket & Geom.\ $n$ & Empirical $\kappa$ & Blom $\kappa$ & Mean gap & q99 gap \\
\midrule
2--8 & 5 & 1.240 & 1.180 & 0.042 & 0.170 \\
8--32 & 21 & 1.945 & 1.890 & 0.083 & 0.292 \\
32--128 & 85 & 2.533 & 2.441 & 0.124 & 0.424 \\
128--512 & 341 & 3.015 & 2.906 & 0.195 & 0.528 \\
\bottomrule
\end{tabular}
\end{table}

Table~\ref{tab:blom-calibration} reports the order-statistic coefficient
$\kappa$ measured on cached internal tiles against Blom's
$\Phi^{-1}((n-0.375)/(n+0.25))$, and the gap between a tile's own cosine and
its subtree maximum (Fig.~\ref{fig:order-stats}). It shows $\kappa$ within
0.06--0.11 of Blom in every bucket (pairwise correlation 0.992). The
99th-percentile gap grows from 0.17 to 0.53 with subtree size, and this
gap is the quantity the 0.99-pinball head is trained to predict. We separate the mechanism by using Blom's formula directly in one experiment, as well as holding n to one in a following experiment. Replacing the learned head with Blom's
formula and one fitted scalar recovers only 5.2 of the 16.4-point R@10 gap
between a fixed-$\kappa$ variance bound and the size-aware head at 5\%
encode. Holding the fitted weights fixed and corrupting only $n$ at
inference costs 11.4 R@10 at 5\% when $n$ is permuted within zoom and 11.8
when $n$ is clamped to one. A per-bucket q99 constant lift trails the
learned head by 11.4.

\subsection{Evaluation protocol details}
\label{supp:protocol}

\paragraph{Image-track site-blocked folds.}
A site is one complete quadtree. For each of three seeds the 256 roots are split into three folds (about 85 test sites; the remaining 171 split 0.2/0.8
into about 34 validation and 137 training sites). Every query ranks the full gallery, and a head is fit only on tiles of its training sites. Stopping margin $\tau$ is selected per fold.

\paragraph{Image-track bound heads.}
$h_\theta$ is $\mathrm{Linear}(2d{+}1,256)\to\mathrm{ReLU}\to\mathrm{Linear}(256,1)$
over $[\mathbf{q},\mathbf{t},\log n]$ with $d{=}1024$. The scalar bound
drops $\log n$. Both are fit with Adam (learning rate $10^{-3}$) for 400
steps of 2{,}048 positive (query, strict-ancestor) rows plus two uniformly
drawn internal training tiles per positive, with the pinball loss of Eq.~4
of the main paper at $\alpha{=}0.99$.

\paragraph{Sampled targets and prediction error.}
Training approximates $c^*$ by a sampled maximum $\widetilde c^*$.
For each internal training tile, the sampler selects $M{=}24$ members
uniformly without replacement from its inclusive subtree when at least
$M$ members are available. The tile itself is eligible, but is not forced
into the sample. Smaller subtrees use all members, with repeated entries
only to pad the tensor to width $M$; repetition does not change the
maximum. The sampled member set is fixed once per fit and used with the
cached query embeddings during optimization. Thus
$\widetilde c^*\leq c^*$, with equality for subtrees containing at most
$M$ members. The cap limits target construction and training-tensor size;
it is a computational choice, not an established optimal sample count.

The sampled-target audit reports underestimation on 32.5\% of positive
query--ancestor rows, with mean shortfall 0.023 over all positive rows,
mostly for the largest subtrees. This measures
$c^*-\widetilde c^*$ before fitting a head. It is distinct from a learned
head's error $c^*-\hat c$: the pinball objective encourages optimistic
predictions relative to the sampled labels, but does not guarantee an
upper bound on the exact subtree maximum. In the scalar-head
training-loss comparison, L2 regression to the same sampled target covers
53\% of training pairs against 99\% for pinball.

A separate held-out diagnostic of the \emph{scalar pinball head}, without
the subtree-size input, reports underestimation on 0.5\% of all
query--tile pairs and 42\% of pairs on the target's ancestor chain,
including the target tile itself. The mean and 99th-percentile shortfalls
over all violating query--tile pairs are 0.030 and 0.168, respectively.
These values characterize that scalar-head diagnostic, not the
size-aware head. Neither aggregate violation rates nor shortfall
quantiles alone certify an adaptive stopping rule: an underestimated
frontier node can contribute to premature stopping before its target is
visited.
The sampling cap has not been swept jointly with head fitting and
adaptive stopping, so these diagnostics do not establish that $M{=}24$
is optimal or quantify how changing $M$ would change stopping recall.
Adaptive recall is measured, not guaranteed.

Three further arms appear in Table~1 of the main paper. \emph{Variance} is
$\hat\mu+\kappa\hat\sigma$ with one constant $\kappa$ per fold, selected on
validation R@10 at 5\% encode from $\{-1,0,0.5,\dots,3\}$. \emph{Node
zoom} is the same MLP with absolute zoom in place of $\log n$. The ZoomEye
schedule blends z-scored cosine, weighted by depth$^2$/max-depth$^2$, with
the z-scored scalar bound, with the blend bias selected from
$\{0,0.2,\dots,1.0\}$ (0 is chosen).

\paragraph{Adaptive stop.}
The stop fires when the best frontier bound falls below the best visited
cosine minus $\tau$. Per fold, $\tau$ is the smallest value on a 0.025 grid
over 0.05--0.60 that reaches 95\% validation reach (selected 0.075--0.20),
with a 60\% encode cap. The deployed stop encodes 6.14\% on average with 93.91\% reach. Ranked by cosine
alone it trails exhaustive cosine ($-2.93$ R@1, $-5.24$ R@10, paired).
MaxSim over the same visited set closes the R@10 gap
(Table~\ref{tab:main-full}). The same rule selects $\tau$ of 0.25--0.30
for the scalar bound (11.11\% mean encode) and needs 25.16\% for raw
cosine. Anchoring the margin to the fifth or tenth best visited cosine
instead of the best raises mean encode to about 10.5\% without raising
reach. A learned stop classifier spans only 2.0--3.7\% mean encode across
its thresholds, so neither replaces the margin rule. Driven by the exact
subtree maximum, the rule stops at 1.47\% mean encode with 97.27 reach and
61.96/86.13/92.51 R@1/5/10 (Table~\ref{tab:main-full}). At $\tau{=}0$ it
stops at 1.30\%, just above the 1.24\% root floor. By then no unencoded
tile can outscore the best tile already found, so its top-1 result is the
exhaustive top-1 (R@1 61.96), but it visits the target on only 71.36\% of
queries because it stops before reaching targets that are not the top-1
tile. The deployed stop's extra cost therefore comes from estimator error,
not from the stop rule.

\paragraph{Intervals.}
Every interval is a 95\% site-block bootstrap: 1{,}000 resamples of
complete site trees over paired deltas. It conditions on the fitted heads and the single query draw, and
stratum intervals are unadjusted for multiplicity. At 10\% encode the
size-aware bound gains 5.60 \ci{+3.06,+8.65} R@10 over zoom-for-$n$, 8.86
\ci{+5.73,+12.23} over the variance bound, and 5.41 \ci{+2.95,+8.53} over
the ZoomEye schedule. Over the scalar bound it gains 12.76
\ci{+7.77,+17.35} at 5\% and 6.31 \ci{+3.52,+9.86} at 10\%.

\paragraph{Cache replay.}
Quality tables replay cached embeddings. A tile's priority depends only on
the query, its own embedding, and its subtree size, so replay reproduces
the on-demand visit order, and cost is charged only on visit
(\S\ref{supp:cost}). The timed runs encode on demand. At full budget the
descent's cosine ranking is identical to exhaustive cosine, with zero
per-query rank differences. The exhaustive late-interaction ceiling is
unchanged by its cosine top-200 shortlist (86.38/97.30/98.88, against
86.38/97.24/98.84 when every token of every tile is scored).

\section{Adapter recipe}
\label{supp:adapter-recipe}

\paragraph{Objective.}
The adapter is trained with a single one-way crop$\to$tile contrastive
loss. A training row is one Git-10M instance: the object is cut from its
anchor tile with the evaluation crop rule and paired with that tile, so the
positive is the anchor, not a parent. A batch of $B$ rows gives $B$ crop
embeddings $\mathbf{q}_i$, $B$ tile embeddings $\mathbf{t}_j$ and a
$B\times B$ logit matrix; the loss for row $i$ is
\begin{equation}
\mathcal{L}_i=-\log
\frac{\exp\bigl(\text{sim}(\mathbf{q}_i,\mathbf{t}_i)/\tau_{\mathrm{c}}\bigr)}
{\sum_{j\in\mathcal{N}(i)}\exp\bigl(\text{sim}(\mathbf{q}_i,\mathbf{t}_j)/\tau_{\mathrm{c}}\bigr)},
\qquad \tau_{\mathrm{c}}=0.03,
\label{eq:supp-i2i}
\end{equation}
where $\mathcal{N}(i)$ is the set of tile columns kept for row $i$: all
columns except those whose tile equals row $i$'s own anchor (two rows can be
cut from one tile). There is no tile$\to$crop term, no hard-negative
mining, and no other loss. A parent or sibling tile that is another row's
anchor is an ordinary negative.

\paragraph{Backbone and adapter.}
Crops and tiles share one encoder, DINOv3 ViT-L/16 pretrained on SAT-493M
(1024-d, no projection head), with every backbone parameter frozen.
Rank-16 LoRA ($\alpha{=}32$, dropout 0.05, $A$ Kaiming-uniform, $B$ zero) is
added to the query, key, value and output projections of every attention
block and both MLP projections (144 modules, 7{,}077{,}888 trainable
parameters, 2.28\% of the backbone). Both sides of Eq.~\ref{eq:supp-i2i}
receive gradient through the shared adapter, and at inference the adapter
is folded into the base weights.

\paragraph{Data and optimisation.}
Rows are Git-10M instances at anchor zooms 10--18 with at least 10 object
pixels. Crop and tile are resized to 224\,px with an independent
horizontal flip (probability 0.5). Optimisation is AdamW at learning rate
$10^{-4}$ and weight decay 0.01, 200 warm-up steps then cosine decay,
batch 96 with two-step gradient accumulation, gradient clipping at 1.0,
mixed precision, and 20 epochs (21{,}940 steps, 6\,h\,54\,min on one
A100). Corrupting only the
query crop (photometric shift, JPEG, rescaling, rotation, sub-window
jitter, padding) changes reach at 25\% encode by at most 2 points.

\paragraph{Training pool and overlap.}
Training instances are the 234{,}058 Git-10M train-split objects with
spatial captions (210{,}622 training, 23{,}436 validation), drawn from the
released pool of over 100 million instances and identity-disjoint from
every Git-10M evaluation query and gallery id. The 5{,}000 Dense500
evaluation objects come from a separate corpus. Geographic footprints can
still intersect: 64 of the 234{,}058 training objects (0.03\%) have anchor
tiles that are ancestors of a Dense500 answer tile. Retraining with those
64 and their quadtree neighbours removed moves none of the 16 size-aware
cells of the budget sweep significantly. The largest shift is 0.95 points,
and the size-aware margins over the scalar bound narrow by at most 1.7
R@10 and stay significant. A change of training seed moves 11 of the 16
cells with a median shift of 1.16, so the overlap matters less than the
seed.

\section{Cost accounting}
\label{supp:cost}

\paragraph{Cost axis.}
Encode cost is backbone forwards on previously unencoded gallery tiles
divided by the gallery size, roots and internal tiles included. The
mandatory roots set a 1.24\% descent floor. Head arithmetic is reported
separately and never substituted for an encoder forward.

\paragraph{Three timing bases.}
\emph{Priced encoder work} is encoded tiles times 2.179\,ms per backbone
forward, measured once on the deployed checkpoint (A100 80GB, batch 64,
TF32). The TFLOPs column is the same tile count times 125.68 GFLOPs.
\emph{Measured wall clock} is the deployed traversal timed on one H100
80GB over 128 queries per budget, with the gallery tensor resident and
query encoding, image preprocessing, and the 11.9\,s gallery preload
excluded. Serial best-first takes a median of 1.72\,s at 5\%, 3.66\,s at
10\%, and 9.49\,s at 25\%. Beam-64 descent takes 1.03, 1.17, and
1.17\,s, and one cold batch-64 pass over the full gallery takes 20.4\,s.
\emph{Decision-only time} excludes the encoder and prices only the
policy's choice of where to look (Table~\ref{tab:policies}; OTM and VAS
batch-amortized, pointer and control timed one query at a time).

\paragraph{Cascades, ANN and amortization.}
A cascade pays a full-gallery cheap pass before its backbone shortlist,
$(N\rho+K)/N$ with $\rho$ measured. The strongest cascade's cheap pass
alone costs 20.8\% of the gallery, so no cascade operates in the 5--15\%
region. ANN indexes speed up lookup over embeddings that already exist and
require a full-gallery encode to build. Pre-encoding costs $N$ forwards
once while 10\% descent costs $0.1N$ per query, so the crossover is about
10 queries per gallery version, and about 20 at 5\%.

These are encoder-work crossovers under the reported no-reuse protocol,
not measured end-to-end latency break-even points. For $Q$ queries to a
fixed gallery and mean encoded fraction $p$, that protocol costs
$QpN$ gallery forwards, whereas a reusable full index costs $N$ forwards
up front, or $N/Q$ amortized forwards per query. This comparison excludes
query encoding, ANN construction and lookup, traversal overhead, and
storage and retrieval of embeddings. Those costs can change the latency
crossover even when the encoder-work counts agree.

Cross-query caching also changes the descent side of this comparison.
With sufficient cache capacity and unchanged imagery and encoder weights,
descent needs to encode only the union of tiles visited across queries;
cache hits require no new gallery forward. That union depends on query
overlap, eviction, and invalidation, and was not measured by our
independent-query evaluation. High query volume over a stable gallery
can amortize pre-encoding, while few queries between gallery or encoder
refreshes can favor encoding on demand. A partial imagery refresh can
invalidate only affected cached tiles, whereas changed encoder weights
can invalidate all embeddings. No workload-level caching or refresh
benchmark is claimed here.

\paragraph{Continental scale.}
The contiguous United States covers 8.08 million km$^2$. A 256-px Web
Mercator tile at zoom $z$ spans $40{,}075\,\mathrm{km}\cdot\cos\varphi/2^{z}$
on a side, about 117\,m at z18 and 40$^\circ$N, so z18 alone holds about
590 million tiles (the ``more than 550 million'' of the main text) and a
complete z10--z20 pyramid about 12.5 billion. At that scale one complete
pre-encode is about 315 A100-days, a 1{,}024-D fp16 pooled index 25.6
decimal TB, and 196$\times$1{,}024 fp16 patch tokens 5.0 decimal PB.
These are storage and encoding-work estimates, not retrieval-quality
measurements at continental scale. Real-imagery retrieval is evaluated on
20{,}608 tiles in 256 site trees. Adding distractor trees can alter frontier
competition, the number of mandatory root encodes, and stopping
calibration; the same encode fractions and recall are not established for
a larger gallery. Section~\ref{supp:synthetic-depth} isolates depth in
synthetic forests. It does not test increasing the number or diversity of
real distractor sites while holding the query set fixed.

\section{Comparison-family diagnostics}
\label{supp:comparison-families}

The object-size buckets of the main paper's Table~3 are the object's
fraction of its tile's area with edges at 1\%, 2\% and 5\% (2{,}582,
1{,}311, 783 and 324 queries); the ``more than 11 R@10'' lead quoted there
is 83.4 against 72.0, paired 11.34 \ci{+6.72,+15.60}.

\subsection{ZoomEye cohort and ranking}
\label{sec:supp:zoomeye}

Table~\ref{tab:zoomeye-harmonized} tests whether ZoomEye's deficit comes from
the tiles it visits or from how it ranks them. ZoomEye runs one traversal per
query to its 25\% cap (the 5\% and 10\% columns are prefixes of that visit
order) and ranks visited tiles by its own existence confidence; re-ranking
the identical tiles by plain cosine, as every other arm does, leaves reach
unchanged and lifts R@10 to within one query of reach at every budget, so
the deficit is almost entirely in which tiles it visits. The Qwen3 cohort
behaves the same way. The 2.8\,h per-query figure of the main text is the
median wall clock over all 100 completed LLaVA and Qwen3 queries on one H100
served by vLLM under concurrency (LLaVA 16, Qwen3 10 concurrent queries),
at least 9{,}584 model forwards per query; the 1.72\,s on the other side of
that comparison is the serial H100 wall clock of \S\ref{supp:cost}.

\begin{table}[htbp]
\centering
\caption{\textbf{ZoomEye (LLaVA-OV-7B), 50-query cohort.} R@10 under ZoomEye's native ranking and under plain-cosine ranking of the same visited tiles.}
\label{tab:zoomeye-harmonized}
\scriptsize
\setlength{\tabcolsep}{5pt}
\begin{tabular}{@{}lrrrrrr@{}}
\toprule
Encode \% & 1.5 & 2 & 3 & 5 & 10 & 25 \\
\midrule
Reach & 2 & 4 & 10 & 12 & 32 & 56 \\
R@10, native confidence ranking & 0 & 0 & 2 & 2 & 2 & 2 \\
R@10, cosine ranking of the same tiles & 2 & 4 & 10 & 12 & 32 & 54 \\
\bottomrule
\end{tabular}
\end{table}

\subsection{Learned policies}

Table~\ref{tab:policies} tests the learned search policies against
\name{} at matched encode. On-policy retraining (OTM) of the scalar bound
head does not improve on its warm start (round-1 R@10 changes of $-3.10$
\ci{-5.69,-0.26} at 5\% and $-3.49$ \ci{-5.86,-1.29} at 10\%), and applied
to the size-aware head it is a null at 5\% and 10\% with small losses at
25\%. The behavior-cloning and
pointwise-control rows use all nine seed--fold cells. The s/query column
excludes the encoder, with OTM and VAS batch-amortized and the pointer and
control rows timed one query at a time (\S\ref{supp:cost}).

\begin{table}[htbp]
\centering
\caption{\textbf{Learned policies versus \name{}.} $\Delta$ is policy minus size-aware R@10 with a paired site-block interval; s/query is decision-only time on one H100.}

\label{tab:policies}
\scriptsize
\setlength{\tabcolsep}{2pt}
\begin{tabular}{@{}lrrrrrr@{}}
\toprule
System & Enc. & Reach & R@1 & R@10 & $\Delta$ R@10 & s/query \\
\midrule
\textbf{\name{}} & 5\% & 89.08 & 55.63 & 83.37 & ref. & --- \\
OTM & 5\% & 74.33 & 46.66 & 69.90 & $-13.47$ \ci{-18.07,-8.50} & 0.0013 \\
VAS & 5\% & 6.22 & 4.24 & 5.97 & $-77.40$ \ci{-81.70,-72.29} & 0.0028 \\
Behavior cloning & 5\% & 53.17 & 33.45 & 49.91 & $-33.46$ \ci{-40.03,-26.62} & 0.223 \\
Pointwise control & 5\% & 53.24 & 33.13 & 49.83 & $-33.53$ \ci{-40.58,-25.96} & 0.016 \\
\midrule
\textbf{\name{}} & 10\% & 96.36 & 59.76 & 89.90 & ref. & --- \\
OTM & 10\% & 89.47 & 55.61 & 83.63 & $-6.27$ \ci{-9.68,-3.41} & 0.0018 \\
VAS & 10\% & 13.33 & 8.57 & 12.75 & $-77.15$ \ci{-83.54,-69.63} & 0.0056 \\
Behavior cloning & 10\% & 72.84 & 44.99 & 67.75 & $-22.15$ \ci{-29.08,-16.03} & 0.524 \\
Pointwise control & 10\% & 72.71 & 44.77 & 67.61 & $-22.29$ \ci{-29.26,-15.70} & 0.037 \\
\midrule
\textbf{\name{}} & 25\% & 99.60 & 61.75 & 92.90 & ref. & --- \\
OTM & 25\% & 98.29 & 60.94 & 91.67 & $-1.23$ \ci{-2.19,-0.59} & 0.0036 \\
VAS & 25\% & 41.27 & 25.61 & 38.65 & $-54.25$ \ci{-65.60,-44.50} & 0.0139 \\
Behavior cloning & 25\% & 89.16 & 55.13 & 83.08 & $-9.82$ \ci{-14.92,-5.58} & 1.462 \\
Pointwise control & 25\% & 91.79 & 56.47 & 85.41 & $-7.49$ \ci{-11.33,-4.53} & 0.119 \\
\bottomrule
\end{tabular}
\end{table}

\subsection{Other search strategies}
\label{supp:other-search}

Table~\ref{tab:mechanisms} tests five other ways of spending the same encode
budget, each driven by the same cached scores so that only the search rule
changes; none reproduces the cited system, each keeps its search rule.
\emph{Per-level beam search} (TDM's native traversal~\cite{zhu2018tdm})
keeps the best $W$ tiles per level before going deeper; it loses to
best-first at every width from 64 up and wins only at 16 and 32, far below
any useful budget. \emph{Greedy single-path} (a proxy for Just Zoom
In~\cite{erzurumlu2026autoregressive}) follows one child per level with no
backtracking and is not significantly different from size-aware best-first
at its 1.3\% encode ($-0.64$ \ci{-1.38,+0.28} R@10). \emph{Attention
sampling}~\cite{katharopoulos2019attention} draws leaves in proportion to a
softmax over their level-1 ancestor's score. \emph{Level-synchronous
top-$K$} (the frontier rule of QuadTree Attention~\cite{tang2022quadtree})
must finish a level before opening the next, so at 5\% and 10\% it reaches
only 4.9\% and 15.4\% of targets and trails size-aware descent by 78.5 and
74.4 R@10. \emph{ZoomSearch-style}~\cite{zhou2025zoomsearch} keeps that
method's sigmoid-threshold branching (top-1 fallback, at most six branches,
depth-dependent cue mix) with our two scores in place of its VLM cues.

\begin{table}[htbp]
\centering
\caption{\textbf{Other search strategies.} Each rule against best-first descent at the same achieved encode: on the same TDM scores for the beam rows, size-aware for the rest. Level-synchronous rows fix the width at 512.}
\label{tab:mechanisms}
\scriptsize
\setlength{\tabcolsep}{3pt}
\begin{tabular}{@{}lrrrr@{}}
\toprule
Mechanism & Enc.\ \% & Reach & R@10 & Best-first R@10 \\
\midrule
Per-level beam, $W{=}16$ & 2.31 & 14.24 & 14.13 & 11.57 \\
Per-level beam, $W{=}64$ & 5.44 & 38.53 & 37.37 & 39.39 \\
Per-level beam, $W{=}128$ & 9.43 & 56.60 & 54.25 & 59.19 \\
Per-level beam, $W{=}512$ & 27.22 & 86.29 & 81.28 & 85.15 \\
\midrule
Greedy single-path & 1.31 & 7.56 & 7.55 & 8.19 \\
Attention sampling & 10.00 & 49.05 & 46.11 & 83.26 \\
Level-synchronous ($K{=}512$) & 5.00 & 4.90 & 4.90 & 83.37 \\
 & 10.00 & 15.45 & 15.43 & 89.86 \\
 & 25.00 & 80.43 & 76.19 & 92.87 \\
ZoomSearch-style schedule & 10.00 & 81.35 & 72.69 & 83.26 \\
\bottomrule
\end{tabular}
\end{table}

\section{Qualitative traversals}
\label{supp:qualitative}

\begin{figure*}[tbp]
\centering
\includegraphics[width=\textwidth]{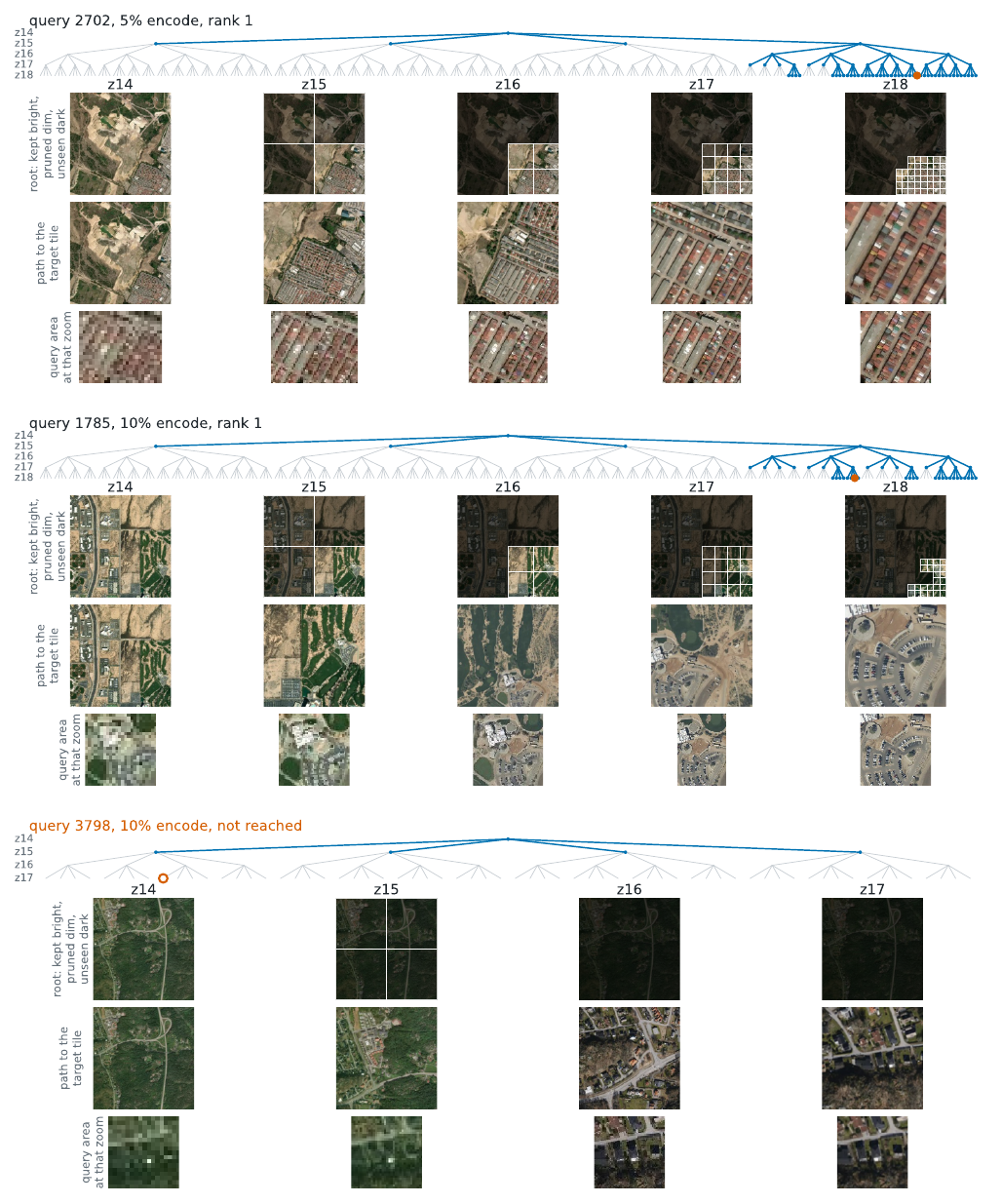}
\caption{\textbf{Three measured descents.} Per case: the site quadtree (blue: encoded; orange: target, hollow if not reached); the root tile per zoom with the tiles the search expanded bright, tiles encoded but not expanded dimmed, and never-encoded tiles dark; the path from root to target; and the query area cut from each path tile. Queries 2702 (5\%) and 1785 (10\%) rank first; 3798 is not reached at 10\%.}
\label{fig:money-traversal}
\end{figure*}

Figure~\ref{fig:money-traversal} shows three measured descents, two that
rank the target first and one that misses it. Each case draws the site
quadtree with the encoded tiles marked, the root tile per zoom level with the tiles the search expanded bright, the tiles it encoded but did not open dimmed, and the rest dark, the path from root to
target, and the query area cut from each path tile. In the miss the site's
root ranked low, its children were encoded only near the end of the budget,
and the correct child was never expanded; the same query is rank~1 at 25\%.

\section{Synthetic depth scaling}
\label{supp:synthetic-depth}

Table~\ref{tab:synthetic-depth} tests how the value of a size-aware
priority changes as pyramids get deeper than Dense500's three to five
levels, on synthetic forests where depth is the controlled variable.
Each forest has sixteen complete 4-ary trees of depth 3--7 (64 to
16{,}384 leaves per tree); tile embeddings follow a hierarchical diffusion on the unit
sphere in 256 dimensions with parent--child cosine $\rho=0.70$, chosen so
that the query--ancestor trail fades into the background cosine within the
tested depths ($0.75\,\rho^{7}\approx 1/\sqrt{256}$); each query mixes one
target leaf with noise to a query--target cosine of 0.75. Search is the
paper's best-first descent with all roots charged, and reach is read from
one traversal at 2\%, 5\% and 10\% of the forest, over 300 queries in each
of three seeds. Raw cosine uses each tile's own similarity. The \emph{size lift} adds a
fixed per-level amount, the 99th percentile of the best-descendant gap on
separate calibration forests, which in a complete tree is the size-aware
bound with the learning replaced by a lookup table. The oracle uses the
exact best-descendant similarity. It shows raw-cosine reach at 5\% falling from 100 to
68.4 as depth goes from 3 to 7 while the size lift stays at or above 99.6
from depth 4 up; at a 2\% budget floor, where even the oracle is starved at
depth 3, the fixed lift over-explores shallow levels and trails cosine at
depths 5 and 6, so a fixed structural prior pays only when the budget
affords the shallow exploration it spends.
This controlled experiment studies a structural mechanism with synthetic
embeddings and a calibrated lookup-table lift. It is not a scaling result
for the trained DINOv3/GeoRSCLIP system on additional real imagery, and
does not establish its recall as the gallery grows wider.

\begin{table}[htbp]
\centering
\caption{\textbf{Synthetic depth scaling.} Reach (\%) of the target leaf by forest depth and budget for raw-cosine and size-lift descent. The oracle reaches 100 in every cell except 74.1 at depth 3 and 2\%.}
\label{tab:synthetic-depth}
\scriptsize
\setlength{\tabcolsep}{3pt}
\begin{tabular}{@{}rrrrrrrr@{}}
\toprule
& & \multicolumn{2}{c}{2\% encode} & \multicolumn{2}{c}{5\% encode} & \multicolumn{2}{c}{10\% encode} \\
\cmidrule(lr){3-4}\cmidrule(lr){5-6}\cmidrule(lr){7-8}
Depth & Leaves & Cosine & Size lift & Cosine & Size lift & Cosine & Size lift \\
\midrule
3 & 64 & 73.0 & 62.6 & 100 & 100 & 100 & 100 \\
4 & 256 & 94.8 & 98.0 & 98.0 & 100 & 99.3 & 100 \\
5 & 1,024 & 85.0 & 78.8 & 92.0 & 99.9 & 95.2 & 100 \\
6 & 4,096 & 70.4 & 63.6 & 81.9 & 99.6 & 88.1 & 100 \\
7 & 16,384 & 54.2 & 99.9 & 68.4 & 100 & 77.4 & 100 \\
\bottomrule
\end{tabular}
\end{table}

\section{Additional reproducibility details}
\label{supp:reproducibility-details}

\subsection{Caption generation and text adaptation}
\label{supp:caption-generation-details}

\begin{table}[htbp]
\centering
\caption{\textbf{Query modalities and reported evaluation populations.}
Each query uses its modality's encoder; image and text embeddings are not
fused into a single query.}
\label{tab:query-modalities-details}
\small
\setlength{\tabcolsep}{4pt}
\begin{tabular}{@{}lrl@{}}
\toprule
Query & Count & Encoder \\
\midrule
Image crop & 5{,}000 & DINOv3 ViT-L/16 + LoRA \\
Caption & 4{,}000 & GeoRSCLIP ViT-L/14 + LoRA \\
\bottomrule
\end{tabular}
\end{table}

\begin{table*}[t]
\centering
\caption{\textbf{Training details for the image-query comparison families.}
The OTM row in the main paper's Table~3 reports the warm-start model;
subsequent refits are evaluated separately in the on-policy study.}
\label{tab:baseline-fitting-details}
\small
\setlength{\tabcolsep}{4pt}
\renewcommand{\arraystretch}{1.12}
\begin{tabular}{@{}p{0.18\textwidth}p{0.77\textwidth}@{}}
\toprule
Family & Objective and fitting settings \\
\midrule
TDM-style~\cite{zhu2018tdm} &
A shared 256-hidden-unit MLP on query and tile embeddings predicts a
binary logit. Strict ancestors of the target are positives; four
negatives per positive are sampled at the same level, with accidentally
positive draws relabelled. Frozen tile embeddings replace free per-tile
parameters so the scorer can operate on held-out sites. \\
OTM-style~\cite{zhuo2020otm} &
A 400-update pinball warm start is followed by three rounds of data
collection and 150-update warm-started refits. Learning rate
$10^{-3}$ is unchanged across rounds. Collection uses the updating policy;
rows are weighted by trajectory counts, with no replay-support mixture.
The adaptation retains subtree-value regression in place of the original
recommendation objective. \\
VAS-style~\cite{sarkar2022vas} &
The flat-action adaptation uses target-hit reward, a training horizon of
64 actions, 30 training epochs, learning rate $3\times10^{-4}$,
32-dimensional keys, and a 64-unit hidden representation. It uses masked
sequential actions and a REINFORCE objective with a random-action control
variate; the gallery is the multiscale tile set rather than a fixed image
grid. \\
Behavior cloning and pointwise control &
Both fit oracle-action demonstrations with AdamW, learning rate
$3\times10^{-4}$, weight decay $10^{-4}$, and logical batches of 256
states. Training runs for at most 30 epochs, with patience five and
selection by validation oracle-action negative log likelihood. Policy
rollouts use the training-site subforest; the pointwise control uses the
same demonstrations and fitting schedule. \\
\bottomrule
\end{tabular}
\end{table*}

The v8 caption-generation pipeline uses InternVL3-38B through
\texttt{lmdeploy}, with a maximum of 130 generated tokens, temperature
$10^{-6}$, top-$p=1$, and top-$k=1$ (greedy decoding). The prompt requests a
35--50-word retrieval description, with the most distinguishing detail in
the first sentence to reduce the effect of subsequent text truncation.
It asks for visible shape, colour, material, relative size, counts, and
spatial arrangement; it discourages generic filler, uncertainty phrases,
and inferred names, history, or interior use. The selected
\texttt{v8\_terse\_exemplars} prompt combines these instructions with four
short examples. Settlement class, land-cover summary, and OSM tags are
provided as auxiliary context to the captioner. The retrieval input is the
generated caption, without appending the context fields, although the
description can carry semantics supplied by that context.

The text encoder is GeoRSCLIP ViT-L/14 with rank-16 LoRA, scaling
$\alpha=32$, on the attention query/key/value/output and MLP projections
of both towers. The backbone weights remain frozen; the adapters and logit
scale are trained. The Git-10M recipe uses eight epochs, batch size 48,
AdamW with learning rate $2\times10^{-4}$ and weight decay $0.01$, and
224-pixel tile inputs. Its objective is the mean of text-to-tile and
tile-to-text cross-entropy losses on normalized embeddings. Off-diagonal
pairs that share the same anchor tile are masked so they are not treated
as negatives. The resulting representation is held fixed during search
policy training and evaluation.

\subsection{Baseline fitting and task adaptation}
\label{supp:baseline-fitting-details}

The embedding-based image-query comparison policies use the shared, frozen image
representation described in \S\ref{supp:adapter-recipe}; policy fitting
does not train a separate image backbone for each method. The scalar and
size-aware bound fitting recipe is given in \S\ref{supp:protocol}.
Table~\ref{tab:baseline-fitting-details} supplies additional settings for
the learned comparison families. These methods operate on the fixed
spatial forest and anchor-tile retrieval objective. The behavior-cloned
pointer uses oracle-action imitation inspired by learned
search~\cite{he2014learningtosearch}; it is not a reproduction of that
work's branch-and-bound solver or DAgger training procedure.

For the text-track OTM study, the historical recipe uses pinball quantile
$0.9$, a 400-update warm start and three 150-update on-policy refits at
learning rate $10^{-3}$. It likewise uses the updating policy,
trajectory-count weights, and no replay-support mixture. These
text-track settings are distinct from the image bound's $0.99$ quantile.

\subsection{Full ZoomEye query and stopping adaptation}
\label{supp:zoomeye-task-adaptation}

The full ZoomEye~\cite{shen2025zoomeye} cohorts in
\S\ref{sec:supp:zoomeye} use image queries. Each backbone first converts
the reference crop to a concise noun phrase describing its principal
object, excluding background, location, and uncertainty. This generation
uses temperature zero and at most 32 new tokens. The cue is produced from
the crop alone, separately from the metadata-assisted captions used in
the text-query benchmark. It is then inserted into ZoomEye's existence,
latent-search, and answerability prompts.

At a root, the MLLM receives the candidate image and the cue. Below a
root, it receives the root image with a red rectangle marking the
candidate footprint, the candidate tile as a zoomed view, and the cue.
Existence asks whether the described object is visible; latent-search
asks whether further zooming could reveal it. Confidence is obtained
from the Yes/No next-token log probabilities. The traversal combines
existence and latent-search confidence using the depth-dependent
quadratic weighting with bias $0.6$, normalized within each site tree.
Every root is scored initially. Expansion visits up to four present
children of the spatial quadtree, replacing recursive pixel splitting
with the dataset's existing tile hierarchy.

The original answerability stopping event is recorded separately from
the fixed-budget trajectory. Its initial threshold is $0.6$ for
LLaVA-OneVision and $0.95$ for the Qwen3-VL port, with lower threshold zero.
After 12 popped nodes, the threshold is reduced every two further pops:
the decrease sequence is $(0.1,0.1,0.2)$ for LLaVA and
$(0.05,0.1,0.2)$ for Qwen3, repeating the final decrease thereafter.
Reaching the natural stop records a snapshot; traversal continues to the
fixed cap. The 5\% and 10\% results are prefixes of the same 25\%
trajectory. A budget counts unique gallery tiles whose priority has been
scored, including roots and frontier tiles; MLLM forwards are counted
separately because a tile uses existence and latent-search probes, while
popped nodes additionally receive an answerability probe.

Native retrieval ranks all visited tiles by existence confidence.
The cosine-ranking diagnostic changes only their order, leaving the
visited set and reach unchanged. Qwen3-VL-8B is a port of the ZoomEye
algorithm, rather than an official upstream configuration. The
ZoomEye-style schedule rows in the main image/text budget sweeps are
separate mechanism baselines driven by the corresponding embedding
scores; they do not run these MLLM prompts on the text-query cohort.

\raggedbottom
\end{document}